\documentclass[a4paper,fleqn]{cas-dc}

\usepackage{algorithm}
\usepackage{algorithmic}
\usepackage{float}
\usepackage{xcolor}
\usepackage{amsmath,amssymb}
\usepackage{booktabs}
\usepackage{multirow}
\usepackage{caption}
\usepackage{placeins}
\usepackage{comment}
\usepackage[authoryear]{natbib}

\hypersetup{
    colorlinks=true,
    citecolor=[rgb]{0,0,0.75},
    linkcolor=blue,
    urlcolor=blue
}
\begin{document}

\let\WriteBookmarks\relax
\def\floatpagepagefraction{1}
\def\textpagefraction{.001}

\shorttitle{CV-SSMNet}
\shortauthors{Zhang et~al.}

\title[mode=title]{Physics-Aware Complex-Valued State Space Model with Scattering-Prior Feature Modulation for PolSAR Image Classification}

\author[1,4]{Fangyan Zhang}[orcid=0000-0003-0739-0307]
\ead{lucia@nxu.edu.cn}

\author[1]{Fan Zhang}[orcid=0000-0002-2058-2373]
\ead{zhangf@mail.buct.edu.cn}

\author[1]{Shiqi Zhou}[orcid=0009-0004-0975-8771]
\ead{1285565899@qq.com}

\author[2]{Jun Ni}[orcid=0000-0002-7105-8475]
\ead{jun.ni@ynu.edu.cn}

\author[3]{Carlos López-Martínez}[orcid=0000-0002-1366-9446]
\ead{carlos.lopezmartinez@upc.edu}

\author[1]{Qiang Yin}[orcid=0000-0002-8413-4756]
\cormark[1]
\ead{yinq@mail.buct.edu.cn}

\affiliation[1]{
organization={The College of Information Science and Technology, Beijing University of Chemical Technology},
city={Beijing},
country={China}
}

\affiliation[2]{
organization={The School of Information Science and Engineering, Yunnan University},
city={Kunming},
country={China}
}
\affiliation[3]{
organization={The Department of Signal Theory and Communications, Polytechnic University of Catalonia},
city={Barcelona},
country={Spain}
}
\affiliation[4]{
organization={The School of Information and Cyberspace Security, Ningxia University},
city={Yinchuan},
country={China}
}
\cortext[1]{Corresponding author: Qiang Yin.\\
\hspace*{1.2em}This work was supported in part by the National Natural Science Foundation of China under Grant No. 62331026
and in part by the Natural Science Foundation of Shandong Province under Grant No. ZR2024ZD19.}

\begin{abstract}
Polarimetric synthetic aperture radar (PolSAR) image classification is a representative task for
physics-aware GeoAI, where land-cover semantics are closely coupled with electromagnetic scattering
mechanisms. Many existing complex-valued networks can preserve amplitude--phase information, but they
are often limited in long-range spatial dependency modeling and usually incorporate polarimetric
priors only as input-level or shallow auxiliary features. As a result, physical knowledge is
insufficiently used to guide deep feature evolution. To address this issue, this paper proposes
CV-SSMNet, a physics-aware complex-valued state-space network with scattering-aware feature
modulation for PolSAR image classification. The proposed method builds a complex-valued state-space
model (CV-SSM) in the original complex domain to capture long-range spatial dependencies while
preserving polarimetric amplitude--phase coupling. Meanwhile, seven physically meaningful scattering
priors, including $H$, $A$, $\alpha$, $P_s$, $P_d$, $P_v$, and $\mathrm{Span}$, are encoded as
FiLM-style modulation signals to adaptively recalibrate complex-valued representations during feature
evolution. CV-SSMNet further integrates multi-scale complex convolutions, branch-wise CV-SSM
encoding, prior-guided recalibration, and lightweight global context aggregation, enabling physically
guided representation learning from local scattering structures to global spatial context.
Experiments on three L-band benchmark datasets and an additional P-band BIOMASS evaluation
demonstrate that CV-SSMNet achieves competitive accuracy, improved regional consistency, and better
boundary preservation, supporting the effectiveness of embedding polarimetric scattering mechanisms
into complex-valued long-range GeoAI representation learning.
\end{abstract}

\begin{highlights}
\item A physics-aware complex-valued state-space learning framework is developed for GeoAI-oriented PolSAR image classification.
\item Polarimetric scattering priors, including H/A/alpha, Freeman-Durden powers, and Span, are encoded as conditional constraints rather than simple auxiliary inputs.
\item Leakage-free spatial block experiments on L-band and P-band PolSAR datasets verify the effectiveness, robustness, and physical interpretability of the proposed model.
\end{highlights}

\maketitle
\section{Introduction}
Polarimetric Synthetic Aperture Radar (PolSAR) is an active microwave remote sensing modality that can acquire scattering information of ground targets regardless of cloud cover and illumination conditions. From the perspective of physics-aware GeoAI, PolSAR image classification requires not only data-driven representation learning but also the explicit incorporation of electromagnetic scattering knowledge into the feature evolution process. Owing to its sensitivity to surface geometry, dielectric properties, and scattering mechanisms, PolSAR has become an important data source for Earth observation, resource survey, and environmental monitoring. With the continuous improvement of sensor spatial resolution and acquisition capabilities, PolSAR data have rapidly increased in scale, dimensionality, and scene complexity, creating new opportunities for fine-grained land-cover mapping and scene understanding. The recent release of large-scale complex-scene datasets such as AIR-PolSAR-Seg-2.0 further highlights the growing practical demand for robust PolSAR terrain classification under realistic conditions~\cite{ref1}. As a fundamental task in Earth observation, PolSAR image classification plays a vital role in land-cover mapping, agricultural monitoring, change detection, and target recognition~\cite{ref2}. PolSAR image classification is a natural scenario for physics-aware machine learning in GeoAI because land-cover semantics are strongly coupled with electromagnetic scattering mechanisms. Therefore, an effective GeoAI model should not only learn discriminative features from data, but also preserve complex-valued scattering information and use physically meaningful polarimetric descriptors to guide representation learning.

Traditional PolSAR image classification methods mainly rely on scattering-mechanism modeling, polarimetric target decomposition, and statistical distribution modeling~\cite{ref3,ref4,ref5}. Representative physical and statistical descriptors, such as $H/A/\alpha$, Freeman--Durden scattering powers, Span, and Wishart-based models, provide interpretable cues for land-cover recognition~\cite{ref6,ref7,ref8,ref9}. However, these methods usually depend on handcrafted features and predefined assumptions, which limits their ability to capture high-level semantic relationships, multi-scale spatial structures, and long-range contextual dependencies in complex scenes.

Deep learning has been widely introduced to improve PolSAR representation learning~\cite{ref10,ref11}. CNN-based, attention-based, self-supervised, and Transformer-style models have shown strong capability in spatial feature extraction and end-to-end classification~\cite{ref12,ref13,ref14,ref15}. For time-series or multi-temporal PolSAR data, recurrent learning, tensor-GCN, and attention-ViT models have also been explored to capture spatial-temporal dependencies~\cite{ref16,ref17,ref18}. Nevertheless, many deep models operate in the real domain and require splitting or mapping complex-valued PolSAR observations into real-valued representations, which may weaken the inherent amplitude--phase coupling.

To preserve complex-valued information, complex-valued deep neural networks have attracted increasing attention~\cite{ref19}. Recent studies have explored complex-valued diffusion models, hybrid complex-valued networks, and complex-valued contourlet neural networks for PolSAR representation learning~\cite{ref20,ref21,ref22}. These methods improve complex feature modeling, but they are still largely built on convolutional or transformer-style operators, making it challenging to efficiently capture long-range spatial dependencies in complex PolSAR scenes.

Another important direction is to incorporate polarimetric scattering priors into deep networks. Physical descriptors such as $H/A/\alpha$, Span, and scattering powers have been used as auxiliary inputs, parallel branches, or fusion features to enhance the sensitivity of networks to scattering mechanisms~\cite{ref23,ref24,ref25,ref26}. However, most existing methods treat these priors as additional features rather than using them as conditional information to regulate deep feature evolution. Therefore, the interaction between physical priors and complex-valued representations remains insufficiently explored.

Recently, state-space models (SSMs) have emerged as an efficient alternative for long-range sequence modeling with linear complexity. SSM-based PolSAR classification has also been preliminarily investigated to enhance contextual dependency modeling~\cite{ref27}. However, existing designs are mostly real-valued, and dedicated complex-valued SSM architectures for jointly modeling amplitude--phase coupling, regional scattering consistency, and long-range context remain underdeveloped.

In summary, existing PolSAR image classification methods still face three major limitations, as illustrated in \textcolor{blue}{Fig.~1 (a)}:
1) physical priors are usually used as input-level features instead of feature-evolution constraints;
2) amplitude-phase coupling and long-range spatial dependency are not jointly modeled in the complex domain; and
3) existing models lack a unified physics-aware framework that combines local scattering structures, global context, and interpretable polarimetric priors.

\begin{figure}
	\centering
	\includegraphics[width=.9\columnwidth]{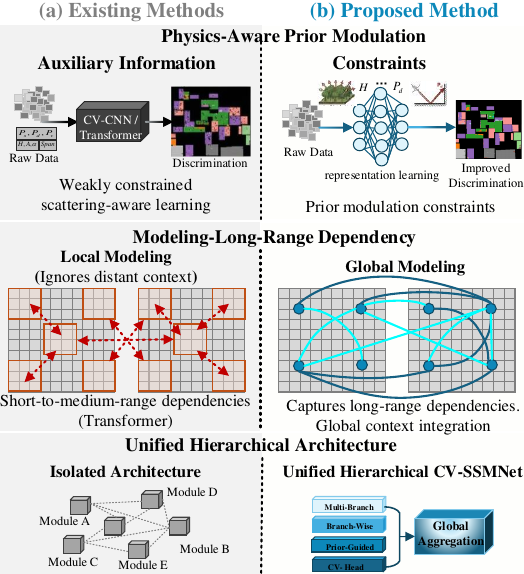}
	\caption{Conceptual comparison between existing PolSAR classification methods (a) and the proposed CV-SSMNet (b) along three dimensions.}
	\label{FIG:1}
\end{figure}

To address these challenges, this paper proposes a prior-guided complex-valued state-space network (CV-SSMNet) for PolSAR image classification. The proposed method uses a state-space model (SSM) as the core operator for efficient
long-range dependency modeling, and extends the state update process to the complex
domain to directly process complex-valued PolSAR scattering features. Furthermore, a scattering-aware prior modulation mechanism is introduced to transform physically meaningful polarimetric priors into conditional signals aligned with the learned representation space, thereby enabling explicit guidance of physical knowledge during complex-valued state-space feature learning. As shown in \textcolor{blue}{Fig.~1 (b)}, the proposed network first extracts local scattering structures using multi-scale complex-valued convolutions, then performs spatial-to-sequence rearrangement and horizontal--vertical bidirectional scanning for long-range context modeling, and finally injects prior modulation and feature enhancement during representation evolution to produce pixel-wise classification outputs.

The main contributions of our method are summarized as follows.

\begin{itemize}
    \item \textbf{Physics-aware scattering-prior-conditioned state-space learning framework:} A polarimetric-prior-conditioned complex state-space learning framework is proposed, where seven polarimetric scattering priors, namely $H$, $A$, $\alpha$, $P_s$, $P_d$, $P_v$, and Span, are transformed from input-side auxiliary information into conditional constraints for the state-space representation learning process. Specifically, conditional representations are generated by a Prior Encoder and injected into complex-valued feature evolution through feature-wise linear modulation (FiLM), scattering-prior-conditioned affine modulation, and prior-guided channel recalibration. In this way, the learned long-range dependencies are driven not only by data statistics but also by physically meaningful scattering mechanisms. In this work, physics-aware learning means that deterministic polarimetric scattering descriptors are transformed into conditional modulation signals that constrain deep complex-valued feature evolution. 

    \item \textbf{Complex-valued state-space modeling paradigm for PolSAR classification:} A complex-valued state-space modeling paradigm is proposed to leverage the linear-complexity long-range dependency modeling capability of SSMs. By means of complex-valued state updates, horizontal--vertical bidirectional scanning,
and stabilized state transition design, amplitude--phase coupling, regional scattering consistency, and cross-region contextual relationships are jointly modeled in the complex domain, thereby mitigating the locality limitation of conventional convolution-based methods.

    \item \textbf{Hierarchical CV-SSMNet collaborative architecture:} A hierarchical architecture is developed to unify local features, multi-scale context, global dependencies, and physical priors. By integrating three-branch multi-scale complex-valued convolutions, branch-wise CV-SSM modules, prior-guided fusion, and a lightweight global CV-SSM module, local scattering structures, multi-scale context, long-range global dependencies, and physics-based prior knowledge are jointly modeled.
\end{itemize}

Furthermore, given the inherent strong spatial autocorrelation of PolSAR imagery, pixel-wise random partitioning may assign neighboring pixels to different subsets, which can introduce spatial information leakage and lead to overly optimistic classification results. Although such random sampling protocols have been widely used in PolSAR image classification, they may be less suitable for evaluating cross-region performance in highly autocorrelated scenes. Therefore, all experiments in this study adopt a spatial block-based partitioning strategy to reduce spatial information leakage between training and test regions. 
\section{Related Work}
\subsection{State Space Models}

State space models (SSMs)~\cite{ref28} provide an efficient framework for long-range sequence modeling through recurrent state updates and convolutional inference. Mamba, introduced by~\cite{ref29}, further introduces data-dependent selective scanning, enabling linear-time modeling of long sequences with improved computational efficiency. Compared with self-attention, SSMs offer a favorable balance between global dependency modeling and scalability.

Recent visual SSMs, such as Vision Mamba~\cite{ref31} and VMamba~\cite{ref32}, extend selective scanning to two-dimensional visual inputs through bidirectional or multi-path scan strategies. In remote sensing, MSFMamba~\cite{ref33} further explores Mamba-based modeling for multi-source image classification by designing multi-scale spatial, spectral, and fusion Mamba blocks, demonstrating the potential of SSMs for heterogeneous remote sensing feature representation and fusion. 

For PolSAR classification, however, the complex-valued nature of the data introduces additional challenges, including amplitude--phase coupling, scattering-mechanism preservation, and stable complex-valued state evolution. Existing visual and remote-sensing SSMs are mainly designed for real-valued visual or multi-source representations, which motivates the development of a complex-valued SSM with scattering-prior-conditioned feature modulation for PolSAR representation learning.

\subsection{Physics-Aware and Prior-Guided Learning in GeoAI}
Recent studies have increasingly explored the incorporation of physical priors into remote sensing representation learning to enhance both discriminability and interpretability. In PolSAR analysis, polarimetric decomposition and scattering-characteristic modeling provide explicit physical cues related to surface scattering, double-bounce scattering, volume scattering, and randomness-dominated scattering mechanisms~\cite{ref34,ref35,ref36}. Such priors are especially beneficial for distinguishing categories that exhibit similar visual appearances but differ in their microwave scattering responses. Existing prior-guided PolSAR methods typically exploit physical descriptors through direct concatenation, parallel feature branches, or late fusion. Although these strategies can improve classification performance, they generally lack an explicit mechanism to regulate the evolution of intermediate representations. In contrast, the proposed method regards polarimetric decomposition features as physics-aware conditioning variables and injects them into the complex-valued state-space learning process via bounded FiLM modulation and prior-guided recalibration, thereby enabling physically informed feature refinement throughout representation learning.

Learning-based methods further exploit such physical information through feature fusion, attention interaction, knowledge-guided learning, and conditional modeling~\cite{ref37,ref38,ref39,ref40}. Instead of relying only on data-driven spatial features, these methods show that polarimetric and scattering priors can guide networks toward mechanism-consistent representations. However, direct concatenation or late fusion may treat physical priors as ordinary auxiliary channels, limiting their ability to adaptively regulate intermediate feature evolution.

Conditional feature modulation provides a more flexible way to inject prior information by generating feature-wise scale and shift parameters from conditioning variables~\cite{ref41}. Motivated by this idea, the proposed SFM formulates the seven polarimetric priors, including $H$, $A$, $\alpha$, $P_s$, $P_d$, $P_v$, and Span, as scattering-aware conditioning variables for complex-valued feature modulation. Different from prompt-learning methods that introduce learnable prompt tokens or rely on pre-trained foundation models, SFM is implemented as a FiLM-style scattering-prior-conditioned modulation module tailored for complex-valued PolSAR representation learning.

\section{Methodology}
\subsection{PolSAR Data Processing}

For a monostatic PolSAR system, the scattering matrix is denoted as
\begin{equation}
\boldsymbol{S}=
\begin{bmatrix}
S_{HH} & S_{HV}\\
S_{VH} & S_{VV}
\end{bmatrix},
\end{equation}
where $H$ and $V$ represent horizontal and vertical polarizations, respectively. Under the reciprocity assumption, $S_{HV}=S_{VH}$. The scattering matrix is then represented in the Pauli basis as
\begin{equation}
\boldsymbol{k}
=
\frac{1}{\sqrt{2}}
\left[
S_{HH}+S_{VV},\,
S_{HH}-S_{VV},\,
2S_{HV}
\right]^{T}.
\end{equation}
Based on the Pauli scattering vector, the multilook coherency matrix is computed as
\begin{equation}
\boldsymbol{T}
=
\frac{1}{L}\sum_{i=1}^{L}
\boldsymbol{k}_{i}\boldsymbol{k}_{i}^{H},
\end{equation}
where $L$ is the number of looks and the superscript $H$ denotes the conjugate transpose.

Since $\boldsymbol{T}$ is Hermitian, its diagonal elements are real-valued and its off-diagonal elements are complex-valued. Therefore, the six upper-triangular elements
$\{T_{11},T_{12},T_{13},\\ T_{22},T_{23},T_{33}\}$ are used as the complex-valued input channels. For each center pixel, a local neighborhood with a size of $13\times13$ is extracted, resulting in an input tensor of size $13\times13\times6$. The patch representation preserves both polarimetric scattering information and local spatial context.

\begin{figure}
	\centering
	\includegraphics[width=.9\columnwidth]{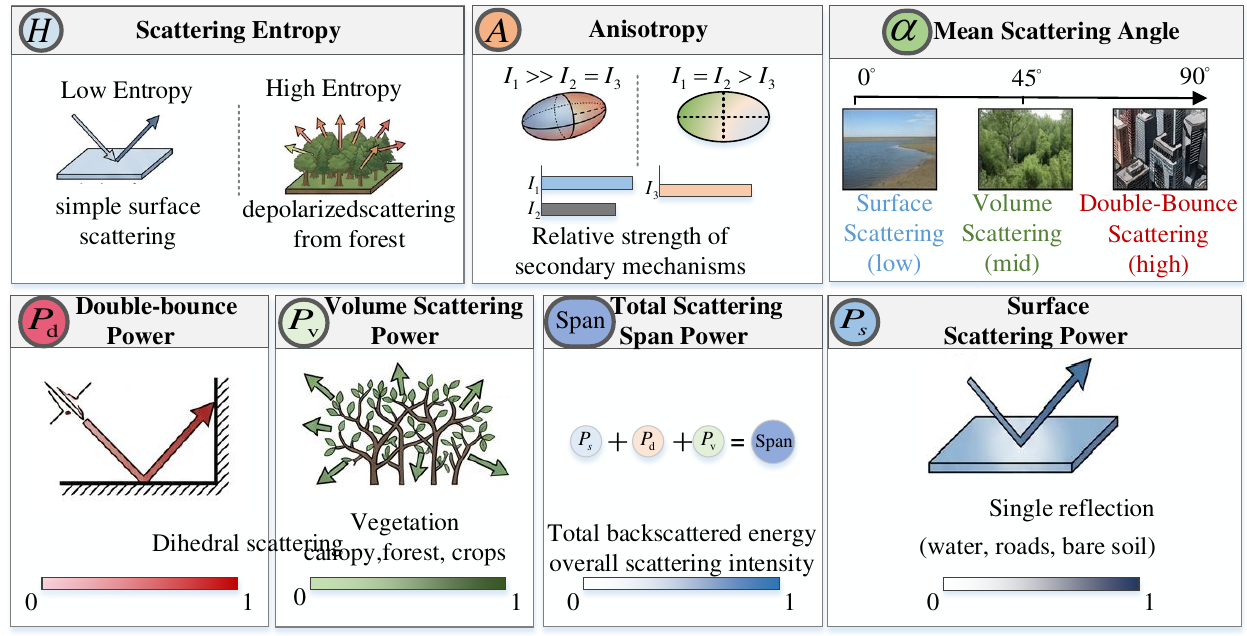}
	\caption{
Illustration of the seven polarimetric scattering priors used in CV-SSMNet,
including entropy $H$, anisotropy $A$, mean scattering angle $\alpha$,
surface scattering power $P_s$, double-bounce scattering power $P_d$,
volume scattering power $P_v$, and total scattering power $\mathrm{Span}$.
}
	\label{fig:2}
\end{figure}

To avoid information leakage during preprocessing, the channel-wise mean and standard deviation are computed only from the training set and then applied to the test set. In addition to the complex-valued coherency-matrix input, seven physically interpretable polarimetric descriptors are used as prior information, including the Cloude--Pottier parameters $H$, $A$, and $\alpha$, the Freeman--Durden scattering powers $P_s$, $P_d$, and $P_v$, and the total scattering power Span. The physical meanings of these seven scattering priors are summarized in \textcolor{blue}{Fig.~2}. These descriptors form the physical prior vector
\begin{equation}
\boldsymbol{p}
=
[H,A,\alpha,P_s,P_d,P_v,\mathrm{Span}].
\end{equation}

\subsection{Overall framework}

\begin{figure*}[!t]
\centering
\captionsetup{justification=raggedright,singlelinecheck=false}
\includegraphics[width=\textwidth]{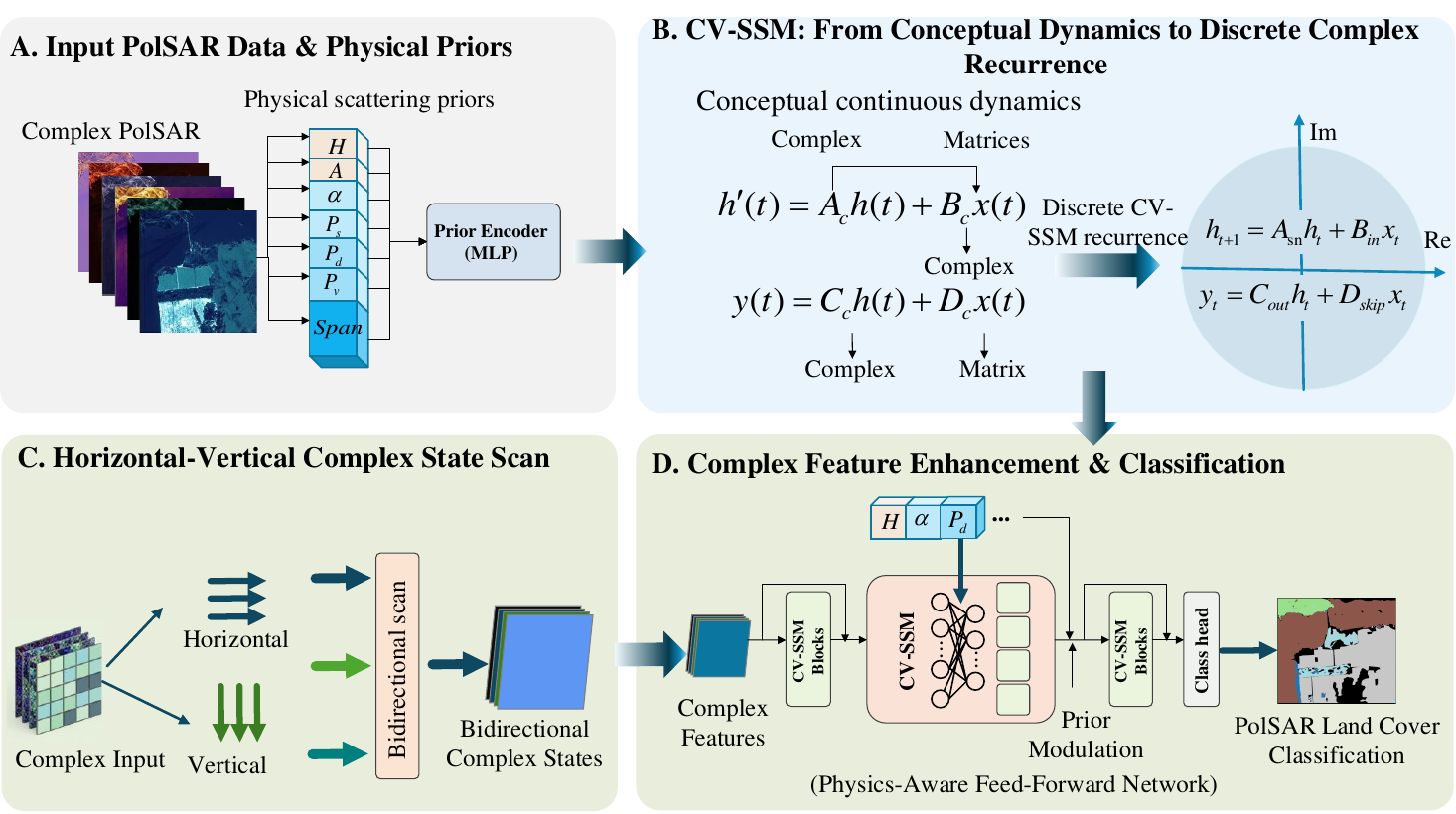}
\caption{
Overall architecture of CV-SSMNet. Complex PolSAR inputs and physical priors
$\{H,A,\alpha,P_s,P_d,P_v,\mathrm{Span}\}$ guide scattering-aware feature
modulation. The CV-SSM illustrates conceptual continuous dynamics and implements
a discrete complex recurrence with $A_{\mathrm{sn}}$, $B_{\mathrm{in}}$,
$C_{\mathrm{out}}$, and $D_{\mathrm{skip}}$, while the subscript $c$ in
$A_c$, $B_c$, $C_c$, and $D_c$ denotes conceptual continuous-domain parameters.
}
\label{fig:3}
\end{figure*}

We first describe the input representation and then introduce the overall pipeline of the proposed method. Our proposed CV-SSMNet consists of two key components: a multi-scale CV-SSM and scattering-aware feature modulation. As illustrated in \textcolor{blue}{Fig.~3}, CV-SSMNet is organized into four functional parts:
input PolSAR data and physical prior encoding, conceptual-to-discrete CV-SSM
modeling, horizontal--vertical complex state scanning, and prior-modulated
feature enhancement for classification.

The input of CV-SSMNet consists of a $13\times13\times6$ complex-valued PolSAR patch
and a seven-dimensional physical prior vector
$\boldsymbol{p}=[H,A,\alpha,P_s,P_d,P_v,\mathrm{Span}]^\top$. The six complex channels are formed by the upper-triangular elements of
the coherency matrix $\boldsymbol{T}$, which preserves polarimetric amplitude,
phase, and amplitude--phase coupling information. Therefore, CV-SSMNet directly
processes the complex-valued PolSAR patch using complex-valued 3D convolutions,
rather than converting it into a purely real-valued representation.

\subsubsection{Stage 1: Multi-Branch Multi-Scale Complex-Valued Convolution}  
A shallow–medium–deep three-branch ComplexConv3D structure with receptive fields of $3\times3$, $5\times5$, and $7\times7$ extracts multi-scale amplitude–phase coupled features. This transforms local spatial structures such as textures, edges, and point targets into higher-dimensional complex channel representations. The resulting 2D feature maps are spatially rearranged into 1D sequences via row-wise and column-wise scans. Each branch is followed by a CV-SSM module, which efficiently aggregates long-range contextual information. The row and column outputs are averaged and reshaped back to the spatial domain. Finally, the three branches are concatenated along the channel dimension to obtain a $13\times13\times6\times48$ complex-valued feature map.

\subsubsection{Stage 2: Scattering-aware Feature Modulation}  
The physical prior $\boldsymbol{p}$ enters a conditional branch to inject physically interpretable preferences into the backbone features. These priors are not raw measurements but low-dimensional summaries of scattering mechanisms (e.g., water typically exhibits high $P_s$ and low $H$, urban areas show high $P_d$, and forests exhibit high $P_v$ and $H$). The prior vector is encoded via a Multilayer Perceptron(MLP) to obtain a conditional embedding $\boldsymbol{z} = \text{MLP}(\boldsymbol{p})$. 

FiLM performs feature-wise conditional affine modulation by generating scaling and shifting parameters $(\lambda, \beta)$ from $\boldsymbol{z}$:

\begin{equation}
\left\{
\begin {aligned}
\lambda &= f_{\lambda}(z) \\
\beta &= f_{\beta}(z)
\end {aligned}
\right.
\end{equation}

where $f_{\lambda}(\cdot)$ and $f_{\beta}(\cdot)$ denote parameterized mapping functions. Subsequently, the calibrated complex features are directly fed into a lightweight CV-SSM for long-range dependency modeling, while prior-guided SE blocks operate on the magnitude responses of complex channels and broadcast the resulting recalibration weights symmetrically to both real and imaginary parts, so that the complex-valued algebraic structure and phase information are preserved.

Finally, adaptive fusion is achieved through Gate Fusion:
\begin{equation}
Y = \delta \odot \mathrm{SSM}(X) + (1-\delta) \odot \mathcal{M}(X,p)
\end{equation}

where $\odot$ denotes element-wise multiplication, $\delta $ denotes learnable gating coefficients, $\delta  \in \mathbb{R}^C$ and $\mathcal{M}(X,p)$ denotes the scattering-prior-conditioned modulation output. $\operatorname{SSM}(\cdot)$  typically denotes the output features resulting from applying State Space Modeling to the input features, serving to capture long-range dependencies. Here, $X$ denotes the complex-valued feature representation extracted from the coherency matrix $\boldsymbol{T}$ via the backbone network. This mechanism dynamically balances data-driven representation learning and physically informed guidance.

\subsubsection{Stage 3: Classification head}  
The fused complex-valued features are flattened and passed through two complex-valued fully connected layers (128 and 64 units) with Dropout regularization. The final complex logits are converted to real-valued probabilities by applying a modulus operator followed by Softmax, producing the final class probabilities.

Overall, this design preserves the phase-related information throughout the network while incorporating physically interpretable prior knowledge, which helps improve robustness and cross-scene classification performance.

\subsection{Physics-aware conditional modulation}

The proposed SFM is the core physics-aware component of CV-SSMNet. It converts polarimetric scattering priors into bounded scaling and shifting parameters, so that the evolution of complex-valued features is guided by physically interpretable scattering mechanisms. SFM explicitly injects polarimetric domain knowledge into the network by using physical scattering priors as conditioning variables to modulate complex-valued backbone features. As illustrated in \textcolor{blue}{Fig.~4}, this module modulates complex-valued features through adaptive scaling and shifting, enabling effective integration of physically meaningful attributes.

The mechanism consists of two components:

\subsubsection{Scattering-aware Feature Modulation Design}
A seven-dimensional physical prior vector $\boldsymbol{p}\in\mathbb{R}^{B\times 7}$ is first encoded by a Prior Encoder implemented as a MLP, producing a conditional embedding:
\begin{equation}
\boldsymbol{z}=\mathrm{MLP}(\boldsymbol{p})
\end{equation}

Based on $\boldsymbol{z}$, two modulation parameters are generated:
a scaling vector $\lambda$ and a shifting vector $\beta$, both constrained to bounded ranges for stable conditioning.

\subsubsection{Complex Feature Modulation}
Given an input complex-valued feature map $F$, the modulation is applied separately to its real and imaginary components:

\begin{equation}
\left\{
\begin{aligned}
Re\left(F^{\prime}\right) &= \lambda_{r} \odot x_{r} + \beta_{r} \\
Im\left(F^{\prime}\right) &= \lambda_{i} \odot x_{i} + \beta_{i}
\end{aligned}
\right.
\end{equation}

where $x_*$ is a complex-valued feature tensor with dimensions $B \times H \times W \times F$, and $\lambda$ is a channel-level modulation coefficient with dimensions $B \times F$. The resulting $F^{\prime}$ is referred to as the modulated complex feature representation. This adaptive modulation mechanism enables the backbone network to leverage physical priors in a controllable manner, improving interpretability and robustness in complex-valued PolSAR feature learning.

\subsubsection{Scattering-aware Feature Modulation Injection Mechanisms}

The scattering-aware feature modulation explicitly injects land-cover scattering characteristics into the network through conditional feature modulation, thereby guiding the learning of complex-valued feature representations. Specifically, a seven-dimensional physical prior vector 
$[H, A, \alpha, \\ P_s, P_d, P_v, \mathrm{Span}]$, derived from polarimetric decomposition, is employed as input. Each component corresponds to a physically interpretable property, including scattering entropy, anisotropy, mean scattering angle, and the power distribution among different scattering mechanisms.

Within the network architecture, the physical priors are first processed by a Prior Encoder, which performs nonlinear mapping to project the low-dimensional prior vector into a high-dimensional conditional space. This conditional representation is aligned with the channel dimension of the backbone features, enabling effective conditioning. Based on this representation, channel-wise scaling parameters $\lambda$ and bias parameters $\beta$ are generated using the FiLM mechanism.

The generated modulation parameters are applied separately to the real and imaginary components of the complex-valued features, preserving the complete complex-valued structure required for phase recovery. Simultaneously, shared scaling modulation helps maintain the consistency of this complex-valued structure, thus preventing the loss of phase-related information. In selected network layers, a gated fusion strategy is further introduced to dynamically balance the contributions of data-driven features and scattering-aware feature modulation.

Through this design, the scattering-aware feature modulation integrates polarimetric scattering priors into the feature evolution process while preserving the structure of complex-valued representations. This provides a physically motivated conditioning mechanism that can improve feature interpretability and robustness across diverse land-cover scenes.

\begin{figure}
	\centering
	\includegraphics[width=.9\columnwidth]{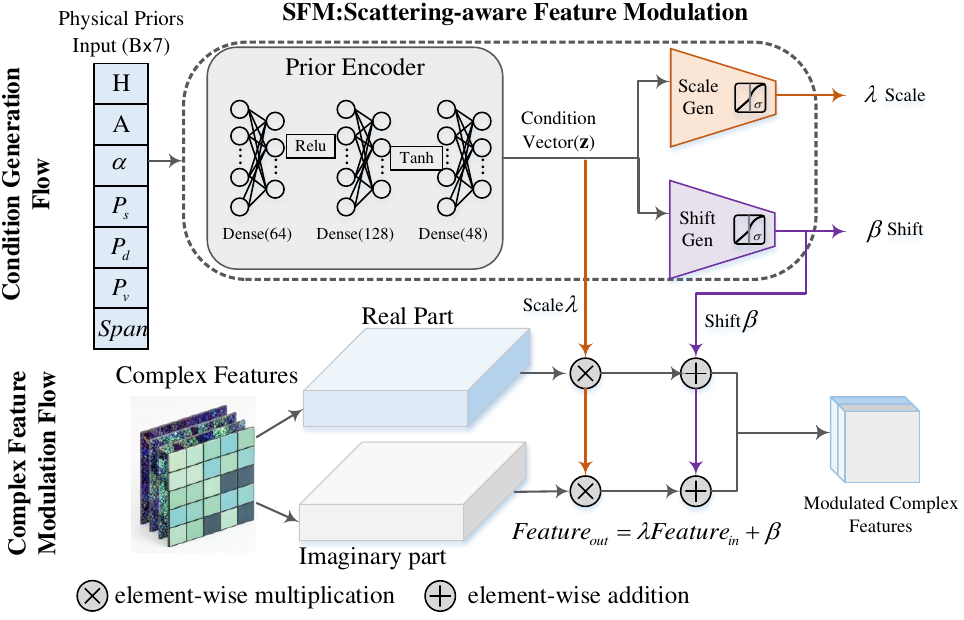}
	\caption{
Structure of the scattering-aware feature modulation (SFM) module. The physical
prior vector is encoded to generate channel-wise scaling $\lambda$ and shifting
$\beta$, which modulate the real and imaginary components of complex features.
}
	\label{Fig:4}
\end{figure}

\subsection{Complex-Valued State Space Model}
CV-SSM is a type of sequential learning framework designed for modeling complex-valued signals, capable of simultaneously characterizing both amplitude and phase information in the complex domain. Compared to traditional real-valued models, CV-SSM is more suitable for processing complex-valued data such as polarimetric SAR data, effectively preserving scattering mechanisms and physical characteristics. Its core idea is to model long-range dependencies through a recursive state-space mechanism. By stabilizing information propagation and global context modeling in the complex feature space, CV-SSM enhances both the expressive power and the robustness of downstream classification.
\subsubsection{Key implementation details of complex-valued SSM}
\textcolor{blue}{Fig.~5} illustrates the implemented discrete CV-SSM structure. 
The complex feature map is first serialized through horizontal and vertical
flattening to form complex sequences, where magnitude and phase information are
preserved. The sequences are then processed by complex state updates, output
projection, and context propagation, enabling stable long-range dependency
modeling in the complex domain.

The input feature map of a CV-SSM block is denoted as
\begin{equation}
\boldsymbol{X} \in \mathbb{C}^{B \times H \times W \times D_p \times C},
\end{equation}
where $B$ is the batch size, $H$ and $W$ are the spatial dimensions, $D_p$ is the polarimetric depth dimension, and $C$ is the feature-channel dimension. Each element of $\boldsymbol{X}$ is a complex value with real and imaginary components. In CV-SSMNet, two sequence construction strategies are used according to the position of the CV-SSM block.

\paragraph{Branch-wise CV-SSM.}
For each multi-scale branch, the feature map is
\begin{equation}
\boldsymbol{X}_b \in \mathbb{C}^{B \times H \times W \times D_p \times C_b},
\end{equation}
where $C_b=16$ in our implementation. In the branch-wise CV-SSM, the polarimetric depth dimension is preserved and folded into the batch dimension during spatial scanning:
\begin{equation}
\boldsymbol{X}_b
\rightarrow
\widetilde{\boldsymbol{X}}_b
\in
\mathbb{C}^{(B D_p) \times (H W) \times C_b}.
\end{equation}
Therefore, the branch-wise CV-SSM uses
\begin{equation}
L_b = H \times W, \quad d_{\mathrm{model}} = C_b.
\end{equation}
For the actual input size $H=W=13$, $D_p=6$, and $C_b=16$, this gives
\begin{equation}
L_b = 13 \times 13 = 169, \quad d_{\mathrm{model}}=16.
\end{equation}

To capture spatial dependencies along both directions, row-wise and column-wise serializations are adopted. For row-wise serialization, the sequence is defined as
\begin{equation}
\left\{
\begin{aligned}
\boldsymbol{S}_{b,d}^{(r)}[t,:] &= \boldsymbol{X}_b[b,h,w,d,:] \in \mathbb{C}^{C_b},\\
t &= h \cdot W + w,
\end{aligned}
\right.
\end{equation}
where $h \in [0,H-1]$, $w \in [0,W-1]$, and $d \in [0,D_p-1]$. Similarly, column-wise serialization is defined as
\begin{equation}
\left\{
\begin{aligned}
\boldsymbol{S}_{b,d}^{(c)}[t,:] &= \boldsymbol{X}_b[b,h,w,d,:] \in \mathbb{C}^{C_b},\\
t &= w \cdot H + h.
\end{aligned}
\right.
\end{equation}
The row-wise and column-wise outputs are averaged and then reshaped back to
$\mathbb{C}^{B \times H \times W \times D_p \times C_b}$.

\paragraph{Lightweight CV-SSM.}
After the three branch-wise outputs are concatenated along the channel dimension, the fused feature map becomes
\begin{equation}
\boldsymbol{X}_f \in \mathbb{C}^{B \times H \times W \times D_p \times C_f},
\end{equation}
where $C_f=48$. In the lightweight CV-SSM, the spatial dimensions and the polarimetric depth dimension are jointly flattened into the sequence dimension:
\begin{equation}
\boldsymbol{X}_f
\rightarrow
\widetilde{\boldsymbol{X}}_f
\in
\mathbb{C}^{B \times (H W D_p) \times C_f}.
\end{equation}
Thus, the lightweight CV-SSM uses
\begin{equation}
L_f = H \times W \times D_p, \quad d_{\mathrm{model}} = C_f.
\end{equation}
For $H=W=13$, $D_p=6$, and $C_f=48$, this gives
\begin{equation}
L_f = 13 \times 13 \times 6 = 1521, \quad d_{\mathrm{model}}=48.
\end{equation}
A single-direction scan with a diagonal state transition matrix is adopted in this lightweight block to reduce computational cost.

For a general complex-valued sequence
$\boldsymbol{S}\in\mathbb{C}^{B' \times L \times d_{\mathrm{model}}}$,
the discrete-time complex SSM is formulated as
\begin{equation}
\left\{
\begin{aligned}
\boldsymbol{h}_{t+1} &= \boldsymbol{A}\boldsymbol{h}_{t}
+ \boldsymbol{B}_{\mathrm{in}}\boldsymbol{x}_{t},\\
\boldsymbol{y}_{t} &= \boldsymbol{C}_{\mathrm{out}}\boldsymbol{h}_{t}
+ \boldsymbol{D}_{\mathrm{skip}}\boldsymbol{x}_{t},
\end{aligned}
\right.
\end{equation}
where $\boldsymbol{x}_{t}\in\mathbb{C}^{d_{\mathrm{model}}}$ is the input token at position $t$, $\boldsymbol{h}_{t}\in\mathbb{C}^{N}$ is the hidden state, and $\boldsymbol{y}_{t}\in\mathbb{C}^{d_{\mathrm{model}}}$ is the output token. The learnable parameters are complex-valued: $\boldsymbol{A}\in\mathbb{C}^{N\times N},\quad$
$\boldsymbol{B}_{\mathrm{in}}\in\mathbb{C}^{N\times d_{\mathrm{model}}},
\quad $
$\boldsymbol{C}_{\mathrm{out}}\in\mathbb{C}^{d_{\mathrm{model}}\times N},
\quad$ $\boldsymbol{D}_{\mathrm{skip}}\in\mathbb{C}^{d_{\mathrm{model}}\times d_{\mathrm{model}}}.$ Here, $\boldsymbol{A}$ controls state evolution, $\boldsymbol{B}_{\mathrm{in}}$ injects the input token into the hidden state, $\boldsymbol{C}_{\mathrm{out}}$ projects the hidden state back to the feature space, and $\boldsymbol{D}_{\mathrm{skip}}$ provides a direct input-output path.

To improve parameter efficiency, the state transition matrix is parameterized
as $\mathbf{A}=\mathbf{A}_0+\mathbf{U}\mathbf{V}^{H}$, where $\mathbf{A}_0$
is a structured base matrix and $\mathbf{U}\mathbf{V}^{H}$ is a low-rank
complex adaptation term. In the lightweight CV-SSM block, a diagonal
$\mathbf{A}_0$ is adopted to reduce computational cost.

The stability of the SSM depends on the spectral radius $\rho(\boldsymbol{A})$. Since
$\rho(\boldsymbol{A}) \leq \sigma_{\max}(\boldsymbol{A})$, spectral normalization is used to constrain the state transition:
\begin{equation}
\boldsymbol{A}_{\mathrm{sn}}
=
\frac{\boldsymbol{A}}
{\max \left(1, \sigma_{\max}(\boldsymbol{A}) / \mu \right)},
\end{equation}
where $\sigma_{\max}(\boldsymbol{A})$ is the largest singular value and $\mu$ controls the upper bound of the spectral norm. In our implementation, $\mu$ is set to $0.95$--$0.98$ for branch-wise CV-SSM and $0.9$ for lightweight CV-SSM. The normalized matrix $\boldsymbol{A}_{\mathrm{sn}}$ is then used in the recurrence:
\begin{equation}
\boldsymbol{h}_{t+1}
=
\boldsymbol{A}_{\mathrm{sn}}\boldsymbol{h}_{t}
+
\boldsymbol{B}_{\mathrm{in}}\boldsymbol{x}_{t}.
\end{equation}

\begin{figure}
	\centering
	\includegraphics[width=.9\columnwidth]{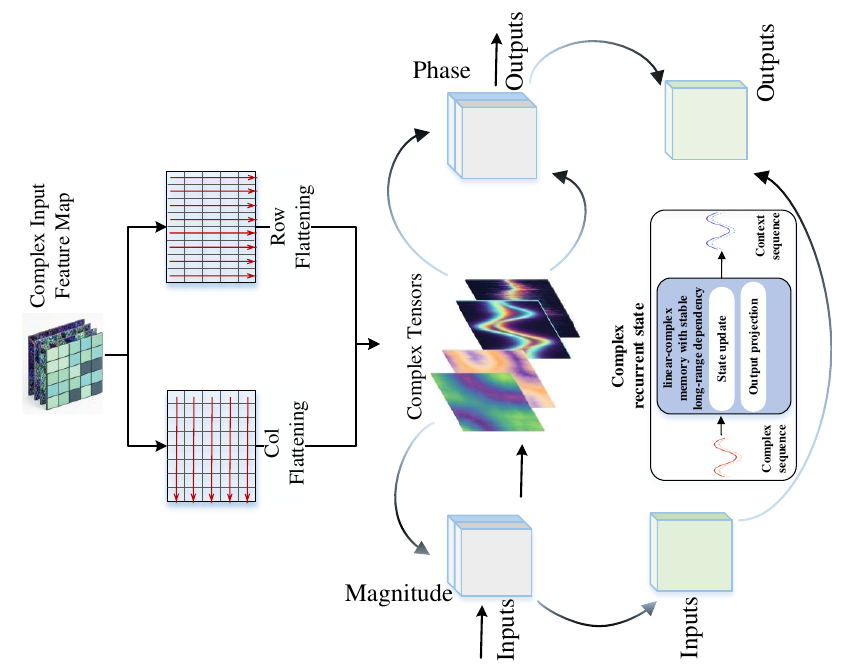}
	\caption{
Structure of the proposed discrete CV-SSM. Row-wise and column-wise
serialization convert complex feature maps into sequences, which are processed
by complex state updates and output projection to model long-range dependencies.
}
	\label{FIG:5}
\end{figure}

Overall, the branch-wise CV-SSM focuses on spatial dependency modeling within each polarimetric depth slice, while the lightweight CV-SSM further captures global dependencies across both spatial positions and polarimetric depth. This clarification explains why the branch-wise CV-SSM uses $L=169$ and $d_{\mathrm{model}}=16$, whereas the lightweight CV-SSM uses $L=1521$ and $d_{\mathrm{model}}=48$.

\subsection{Hierarchical CV-SSMNet Collaborative Architecture}

To address the insufficient use of physical priors, limited long-range modeling in the complex domain, and fragmented representation learning, CV-SSMNet integrates local scattering feature extraction, multi-scale context modeling, complex-valued long-range dependency learning, and physical-prior guidance into a unified end-to-end architecture, as summarized in \textcolor{blue}{Table~\ref{tab:Table0}}.

\subsubsection{Multi-Branch Multi-Scale Complex-Valued Convolution}

PolSAR scenes contain targets with different scales, spatial layouts, and scattering heterogeneity. Therefore, three parallel complex-valued convolution branches with receptive fields of $3 \times 3$, $5 \times 5$, and $7 \times 7$ are used to capture fine details, mesoscale patterns, and broader contextual structures, respectively. All branches operate directly in the complex domain, preserving amplitude--phase coupling and polarimetric correlations while avoiding information loss caused by premature real-valued decomposition.

Given an input complex-valued PolSAR patch
\begin{equation}
X \in \mathbb{C}^{H \times W \times D \times C},
\end{equation}
where $H$ and $W$ denote spatial dimensions, $D$ is the polarimetric/depth-related dimension, and $C$ is the channel number, the three branches generate
\begin{equation}
F^{(1)},\,F^{(2)},\,F^{(3)} \in \mathbb{C}^{H \times W \times D \times C}.
\end{equation}
These multi-scale features preserve local scattering structures and provide complementary representations for subsequent CV-SSM encoding.

\subsubsection{Branch-Wise CV-SSM Encoding}

Although complex-valued convolutions effectively extract local features, their receptive fields are still limited. To capture long-range scattering consistency and contextual dependencies, an independent CV-SSM encoder is assigned to each branch. For each branch, the spatial feature map is rearranged into a sequence and processed by complex-valued state updates with horizontal--vertical bidirectional scanning. This design connects local multi-scale extraction with nonlocal contextual modeling while preserving amplitude--phase coupling.

\begin{table*}[t]
\caption{Layer-wise configuration of CV-SSMNet.}
\label{tab:Table0}
\centering
\footnotesize
\renewcommand{\arraystretch}{1.08}
\setlength{\tabcolsep}{4pt}
\begin{tabular*}{\textwidth}{@{\extracolsep{\fill}}lllll}
\toprule
Stage & Layer & Input & Output & Description \\
\midrule
Input & \texttt{PolSAR\_input} & -- & $(B,13,13,6,1)$ & Complex-valued patch \\
Input & \texttt{prior\_input} & -- & $(B,7)$ & Physical prior vector \\
Multi-scale branches & \texttt{shallow/mid/deep\_conv} 
& $(B,13,13,6,1)$ & $3\times(B,13,13,6,16)$ 
& 1/2/3 CV-Conv3D layers \\
Branch-wise modeling & \texttt{branch\_cvssm} 
& $(B,13,13,6,16)$ & $(B,13,13,6,16)$ 
& CV-SSM per branch \\
Feature fusion & \texttt{features\_concat} 
& $3\times(B,13,13,6,16)$ & $(B,13,13,6,48)$ 
& Channel concatenation \\
Prior modulation & \texttt{concat\_prior\_cond} 
& $(B,13,13,6,48)$ & $(B,13,13,6,48)$ 
& FiLM conditioning \\
Refinement & \texttt{se\_block\_1--3} 
& $(B,13,13,6,48)$ & $(B,13,13,6,48)$ 
& Prior-guided SE \\
Global modeling & \texttt{lightweight\_cvssm} 
& $(B,13,13,6,48)$ & $(B,13,13,6,48)$ 
& Lightweight CV-SSM \\
Classification head & \texttt{flatten+dense} 
& $(B,13,13,6,48)$ & $(B,K)$ 
& Complex dense + magnitude softmax \\
\bottomrule
\end{tabular*}

\vspace{1mm}
\begin{flushleft}
\footnotesize
Note: Branch-wise CV-SSM uses $(B D_p, H W, C_b)$ with $L=169$ and
$d_{\mathrm{model}}=16$; lightweight CV-SSM uses $(B, H W D_p, C_f)$ with
$L=1521$ and $d_{\mathrm{model}}=48$.
\end{flushleft}
\end{table*}

\subsubsection{Prior-Guided Recalibration and Cross-Branch Fusion}

Physical priors are not simply concatenated with image features. Instead, the seven polarimetric priors $H$, $A$, $\alpha$, $P_s$, $P_d$, $P_v$, and $\mathrm{Span}$ are encoded by a Prior Encoder into conditional embeddings aligned with the feature space. These embeddings guide feature-wise linear modulation, scattering-prior-conditioned CV-SSM modulation, and prior-guided channel recalibration, thereby enhancing scattering-consistent responses and suppressing redundant activations.

After branch-wise local-context modeling, the three branches are fused into a unified multi-scale representation and further refined by a lightweight global CV-SSM module. Since scale-specific dependencies have already been modeled in each branch, the global module mainly performs holistic integration, balancing representation capability and computational efficiency. The final representation jointly encodes local scattering structures, multi-scale context, long-range dependencies, and physics-based semantic guidance.

\subsubsection{Complex-Valued Classification Head}

To retain complex-valued representations until classification, a complex-valued fully connected layer is used. Given $z=x_r+jx_i$, $W=W_r+jW_i$, and $b=b_r+jb_i$, the output is
\begin{equation}
y = zW + b,
\end{equation}
which can be expanded as
\begin{equation}
\begin{aligned}
y_r &= x_rW_r - x_iW_i + b_r,\\
y_i &= x_rW_i + x_iW_r + b_i.
\end{aligned}
\end{equation}
Thus, the real and imaginary components jointly contribute to classification. Complex-valued batch normalization, synchronized dropout, and Cartesian complex ReLU are used for stable optimization:
\begin{equation}
f(z)=\mathrm{ReLU}(x_r)+j\,\mathrm{ReLU}(x_i).
\end{equation}

In the prior modulation module, the scaling and shifting parameters are generated from the physical prior vector $\mathbf{p}$:
\begin{equation}
\lambda = 1 + 0.5\tanh(f_\lambda(\mathbf{p})), \quad
\beta = 0.5\tanh(f_\beta(\mathbf{p})).
\end{equation}
The modulation is applied separately to the real and imaginary parts:
\begin{equation}
\begin{aligned}
x_r' &= \lambda_r \odot x_r + \beta_r,\\
x_i' &= \lambda_i \odot x_i + \beta_i.
\end{aligned}
\end{equation}

Finally, complex logits are converted to magnitudes before Softmax. For the $k$-th class logit $y_k=y_{r,k}+jy_{i,k}$,
\begin{equation}
|y_k|=\sqrt{y_{r,k}^2+y_{i,k}^2},
\end{equation}
and the class probability is
\begin{equation}
P_k=\frac{\exp(|y_k|)}{\sum_{j=1}^{K}\exp(|y_j|)},
\end{equation}
where $K$ is the number of classes, $P_k \ge 0$, and $\sum_{k=1}^{K}P_k=1$.

\section{Experiments}
\subsection{Dataset Description}
The proposed method is comprehensively evaluated on four real-world PolSAR datasets covering different sensors, frequency bands, spatial resolutions, and land-cover scenarios. Three widely used L-band benchmarks, namely Flevoland, San Francisco, and Oberpfaffenhofen, are adopted to evaluate classification performance in agricultural, urban, and mixed land-cover scenes. In addition, the P-band ESA BIOMASS dataset is used as a large-scale evaluation scenario to examine the applicability of CV-SSMNet under P-band observations with different scattering characteristics. The Pauli-RGB images and ground-truth maps of the four datasets are shown in \textcolor{blue}{Fig.~6}.

\textbf{1) Flevoland Dataset}~\cite{ref44}: 
The Flevoland dataset was acquired in 1989 over the Flevoland region of the Netherlands by the NASA/JPL AIRSAR sensor. It consists of L-band fully polarimetric SAR data with a spatial resolution of approximately 6 m and contains 15 crop categories, making it a classic benchmark for agricultural PolSAR classification.

\textbf{2) San Francisco Dataset}~\cite{ref45}: 
The San Francisco dataset was collected over the San Francisco Bay area by the AIRSAR airborne sensor. It provides L-band fully polarimetric SAR data with a spatial resolution of about 10 m and includes typical urban and suburban land-cover types, such as buildings, vegetation, and water.

\textbf{3) Oberpfaffenhofen Dataset}~\cite{ref46}: 
The Oberpfaffenhofen dataset was acquired in 2003 over Germany by the DLR E-SAR sensor. It contains L-band fully polarimetric SAR data with a spatial resolution of approximately 3 m, covering built-up areas, roads, open areas, and vegetation.

\textbf{4) BIOMASS Dataset}: 
The BIOMASS dataset is derived from the ESA BIOMASS mission, the first spaceborne fully polarimetric P-band SAR mission for global forest above-ground biomass estimation~\cite{ref47}. Compared with L-band benchmarks, P-band observations provide stronger canopy penetration and different polarimetric scattering statistics, leading to greater spatial variability. Therefore, BIOMASS offers a challenging large-scale P-band test case. Since the model is trained and evaluated within this dataset, this experiment should be regarded as a P-band evaluation rather than a strict cross-band transfer setting.

\begin{figure*}[!t]
\centering
\captionsetup{justification=raggedright,singlelinecheck=false}
\includegraphics[width=\textwidth]{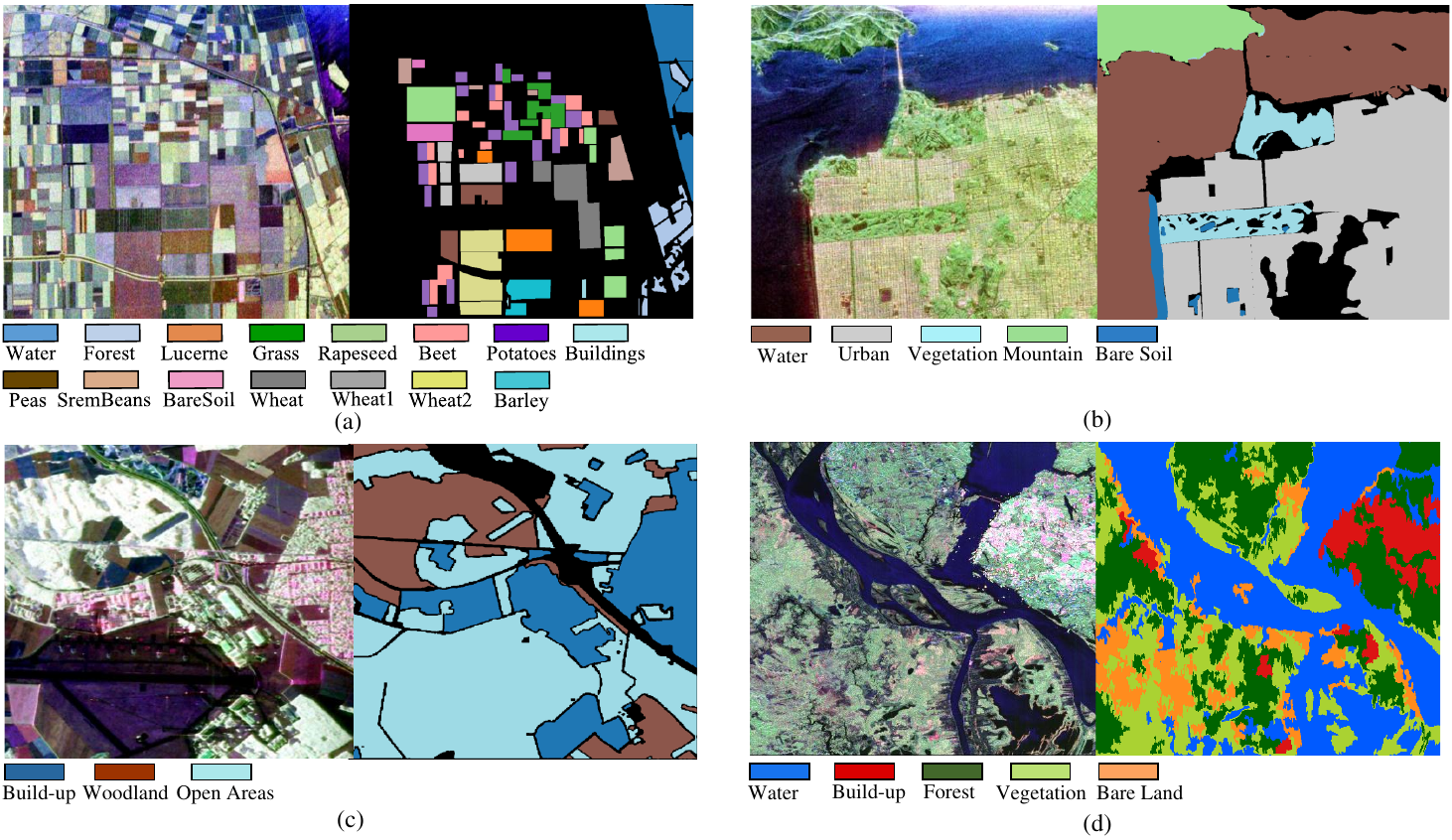}
\caption{Pauli RGB images and ground-truth maps of the four datasets used in this study. (a) Flevoland dataset (b) San Francisco dataset (c) Oberpfaffenhofen dataset (d) BIOMASS dataset.}
\label{fig:6}
\end{figure*}

\subsection{Implementation Details}
\subsubsection{Mitigating Spatial Information Leakage via Block-Based Data Splitting}

\begin{table*}[t]
\centering
\caption{Statistics of the leakage-free spatial split protocol.}
\label{tab:spatial_split_statistics}
\resizebox{\textwidth}{!}{%
\begin{tabular}{lcccccccccc}
\hline
Dataset & Block size & Total labeled & Train & Train (\%) & Test & Test (\%) & Guard & Guard (\%) & Unused & Unused (\%) \\
\hline
Flevoland & $35\times35$ & 207,832 & 2,086 & 1.00 & 172,165 & 82.84 & 24,648 & 11.86 & 8,933 & 4.30 \\
San Francisco & $35\times35$ & 802,302 & 8,026 & 1.00 & 666,700 & 83.10 & 95,660 & 11.92 & 31,916 & 3.98 \\
Oberpfaffenhofen & $35\times35$ & 1,311,618 & 13,117 & 1.00 & 1,085,746 & 82.78 & 157,723 & 12.03 & 55,032 & 4.20 \\
Biomass & $35\times35$ & 199,962 & 2,002 & 1.00 & 167,788 & 83.91 & 23,020 & 11.51 & 7,152 & 3.58 \\
\hline
\end{tabular}%
}
\end{table*}

\begin{figure*}[!t]
\centering
\captionsetup{justification=raggedright,singlelinecheck=false}
\includegraphics[width=\textwidth]{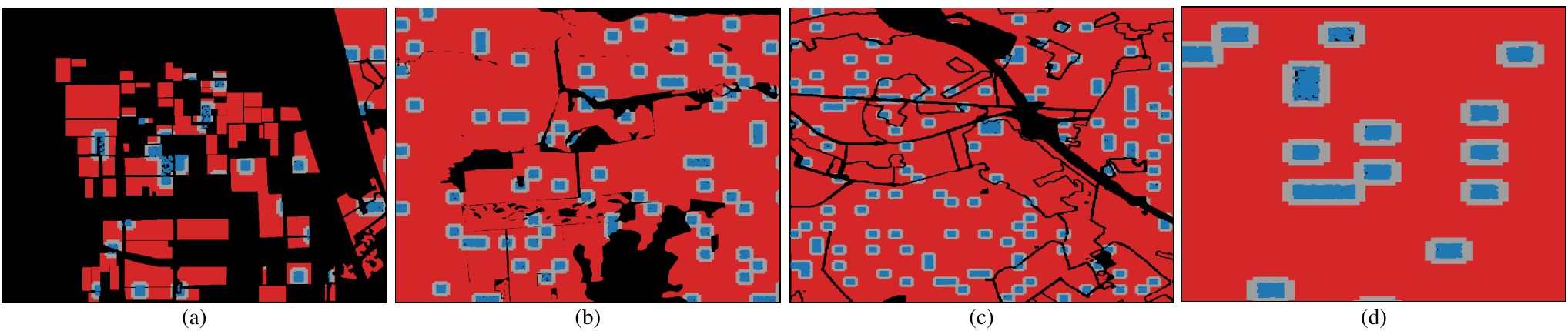}
\caption{Visualization of the leakage-free global spatial block partitioning protocol. 
(a) Flevoland, (b) San Francisco, (c) Oberpfaffenhofen, and (d) Biomass. 
Blue pixels denote the selected 1\% training sample centers, red pixels denote the spatially disjoint test samples, gray regions denote the guard bands discarded to prevent $13\times13$ patch overlap, and black regions denote background or unused labeled pixels.}
\label{fig:7}
\end{figure*}

To ensure a reproducible and leakage-free evaluation, a global spatial block partitioning protocol is adopted for all datasets. As shown in \textcolor{blue}{Fig.~7}, the labeled map is first divided into non-overlapping spatial blocks with a size of $35\times35$. The split is performed globally over the whole scene rather than independently for each class, so that neighboring pixels from different classes are assigned consistently to the same spatial subset. Following the common low-shot PolSAR setting, only 1\% labeled pixels of each class are selected as training sample centers from the training candidate blocks, while the retained pixels in the spatially disjoint test regions are used for evaluation. Since the input patch size is $13\times13$, a 6-pixel guard band, corresponding to the patch radius, is removed along train/test boundaries to prevent contextual overlap between patches from different subsets. Pixels in the guard band and unused pixels in training candidate blocks are excluded from both training and testing. The BIOMASS reference map used for classification has a size of $529\times378$, containing 199,962 labeled pixels. The resulting sample statistics are summarized in \textcolor{blue}{Table~\ref{tab:spatial_split_statistics}}.

\subsubsection{Experimental environment and details}

All experiments were implemented with \texttt{TensorFlow 2.6}, the \texttt{cvnn} library, and \texttt{Python 3.9}, and conducted on an NVIDIA RTX A5500 GPU with 24\,GB memory.

For fair comparison with CV-ASDF2Net~\cite{ref19}, the same 1\% class-wise low-shot training setting is adopted. Test samples are selected using the proposed spatially disjoint block partition rather than pixel-wise random sampling, which helps reduce information leakage caused by spatial autocorrelation. Specifically, each PolSAR image is divided into non-overlapping spatial regions; training samples are drawn from training regions, while labeled pixels in testing regions are used only for evaluation. The class-wise training/testing numbers are reported in Tables~\ref{tab:Table1}--\ref{tab:Table3}. All compared methods and ablation variants use identical spatial partitions, sample numbers, random seeds, and evaluation protocols.

{Implementation Details.}
CV-SSMNet adopts a dual-input setting, including a complex-valued PolSAR patch of size $13\times13\times6$ with shape $(B,13,13,6,1)$ and a 7-D physical prior vector $\mathbf{p}=[H,A,\alpha,P_s,P_d,P_v,\mathrm{Span}]$. The feature extractor contains three multi-scale \texttt{ComplexConv3D} branches with 1/2/3 convolutional layers for shallow, middle, and deep features. Each convolution uses a $3\times3\times3$ kernel, 16 filters, same padding, and \texttt{cart\_relu}. The branch outputs are concatenated into $(B,13,13,6,48)$.

{Complex-valued SSM.}
CV-SSMNet includes three branch-wise CV-SSM blocks and one lightweight CV-SSM block. For each branch-wise CV-SSM, features of shape $(B,13,13, \\ 6,16)$ are reshaped to $(B\cdot6,169,16)$, where $L=169$ and $d_{\mathrm{model}}=16$. For the lightweight CV-SSM, the fused feature $(B,13,13,6,48)$ is reshaped to $(B,1521,48)$, where $L=1521$ and $d_{\mathrm{model}}=48$.

{Scattering-aware Feature Modulation and Stabilization.}
Physical priors are injected through a FiLM conditioner. The prior encoder maps the 7-D prior vector to a $(B,128)$ condition using an MLP $7\!\rightarrow\!64\!\rightarrow\!128\!\rightarrow\!128$ with Tanh activation. The FiLM generator produces channel-wise modulation parameters $\lambda,\beta\in\mathbb{R}^{C}$ with $C=48$. Prior-guided SE blocks and gated fusion enhance scattering-aware feature selection. Spectral normalization constrains the spectral radius of $A$ to 0.95--0.98 for branch-wise CV-SSM blocks and 0.9 for the lightweight block. The classifier uses \texttt{ComplexDense} layers with 128/64 units, \texttt{ComplexDropout} with a rate of 0.25, and a magnitude-based softmax output.

{Training Details.}
The model is optimized by Adam with weight decay $1\times10^{-4}$ and an initial learning rate of $1\times10^{-3}$. A cosine learning-rate schedule with a 5-epoch warmup and global gradient clipping with a max norm of 1.0 are used. The model is trained for 100 epochs with a batch size of 64. Unless otherwise specified, results are reported as mean $\pm$ standard deviation over five independent runs, using OA, AA, kappa coefficient, and per-class accuracy as evaluation metrics.

\subsubsection{Comparison Methods}
To evaluate the effectiveness of the proposed CV-SSMNet, comparisons are conducted with seven representative PolSAR and remote sensing classification methods. These baselines are selected from four perspectives: label-efficient PolSAR representation learning, complex-valued spatial-scattering feature extraction, attention-based contextual modeling, and state-space/Mamba-based long-range dependency modeling.

For fair comparison, all methods are evaluated under the same leakage-free spatial protocol, including identical training sample centers, test regions, guard bands, patch size, data augmentation, batch size, training epochs, and evaluation metrics. Methods with official implementations are retrained using their released code whenever available, while methods without released code are reimplemented according to the original papers. To prevent test data leakage, the spatially disjoint test regions are used only for final evaluation and are never used for training or hyperparameter tuning. Hyperparameters are fixed before final testing based on preliminary experiments conducted only within the training regions. The learning rate, weight decay, and dropout are selected from $\{5\times10^{-4},10^{-3},2\times10^{-3}\}$, $\{10^{-5},10^{-4},5\times10^{-4}\}$, and $\{0,0.25,0.5\}$, respectively.

\begin{itemize}
    \item \textbf{Self-supervised and contrastive PolSAR representation learning:}
    PCLNet~\cite{ref48}, PiCL \\~\cite{ref13}, and SSL-MBC~\cite{ref14} are selected to assess label-efficient PolSAR feature learning.

    \item \textbf{Complex-valued spatial-scattering feature extraction:}
    CV-3DCNN~\cite{ref49} and CV-ASDF2Net \\~\cite{ref19} are adopted to evaluate complex-valued feature modeling and hierarchical scattering representation.

    \item \textbf{Transformer-based contextual modeling:}
    PolSARFormer~\cite{ref15} is included to compare CV-SSMNet with attention-based PolSAR classification.

    \item \textbf{State-space/Mamba-based long-range modeling:}
    MSFMamba~\cite{ref33} is introduced as a recent Mamba-based remote sensing baseline for evaluating efficient long-range dependency modeling.
\end{itemize}

\subsection{Comparison and visualization analysis with SOTA methods}
This section comprehensively verifies the effectiveness of the proposed CV-SSMNet in PolSAR image classification tasks through systematic comparative experiments with seven representative state-of-the-art methods. Quantitative metrics and visualization results analysis evaluate the model's advantages in complex-valued long-range dependency modeling and physical prior guidance.

\textcolor{blue}{Fig.~8} shows the visualization results on the Flevoland dataset, distinct methods exhibit significant differences regarding the regional purity, boundary refinement, and spatial consistency of their classification maps. Comparative methods—such as PCLNet, CV-3DCNN, and MSFMamba achieve reasonable performance in identifying major land-cover categories; however, they still manifest noticeable class mixing and speckle noise within complex boundaries and small-scale target regions. While PiCL, PolSAR-Former, CV-ASDF2Net, and SSL-MBC demonstrate improvements in terms of spatial continuity and regional integrity, they nonetheless remain deficient in the recovery of local details. In contrast, the proposed CV-SSMNet yields classification results that align more closely with the ground truth annotations; it not only effectively suppresses intra-regional noise but also preserves superior structural integrity and boundary clarity at parcel boundaries and within fine-grained target areas, thereby indicating the effectiveness of the proposed method in improving overall
regional consistency and boundary preservation.

As shown in \textcolor{blue}{Table~\ref{tab:Table1}}, CV-SSMNet achieves the best overall performance on the Flevoland dataset, obtaining the highest OA of 97.56\% and Kappa$\times$100 of 97.05. It reaches 100\% accuracy on several classes, including Water, Forest, Grass, Bare Soil, three Wheat-related classes, Barley, and Buildings, demonstrating strong global consistency and reliable discrimination for categories with distinctive scattering responses. Nevertheless, its advantages are not uniform across all classes. For visually similar crop categories, such as Lucerne, Beet, Potatoes, and Peas, the accuracies drop to 86.42\%, 72.97\%, 83.23\%, and 89.30\%, respectively, which are lower than several competing SOTA methods. This also explains why its AA of 95.13\% is slightly lower than SSL-MBC. These results indicate that CV-SSMNet improves overall classification consistency, while fine-grained crop discrimination remains challenging.

\begin{figure*}[!t]
\centering
\captionsetup{justification=raggedright,singlelinecheck=false}
\includegraphics[width=\textwidth]{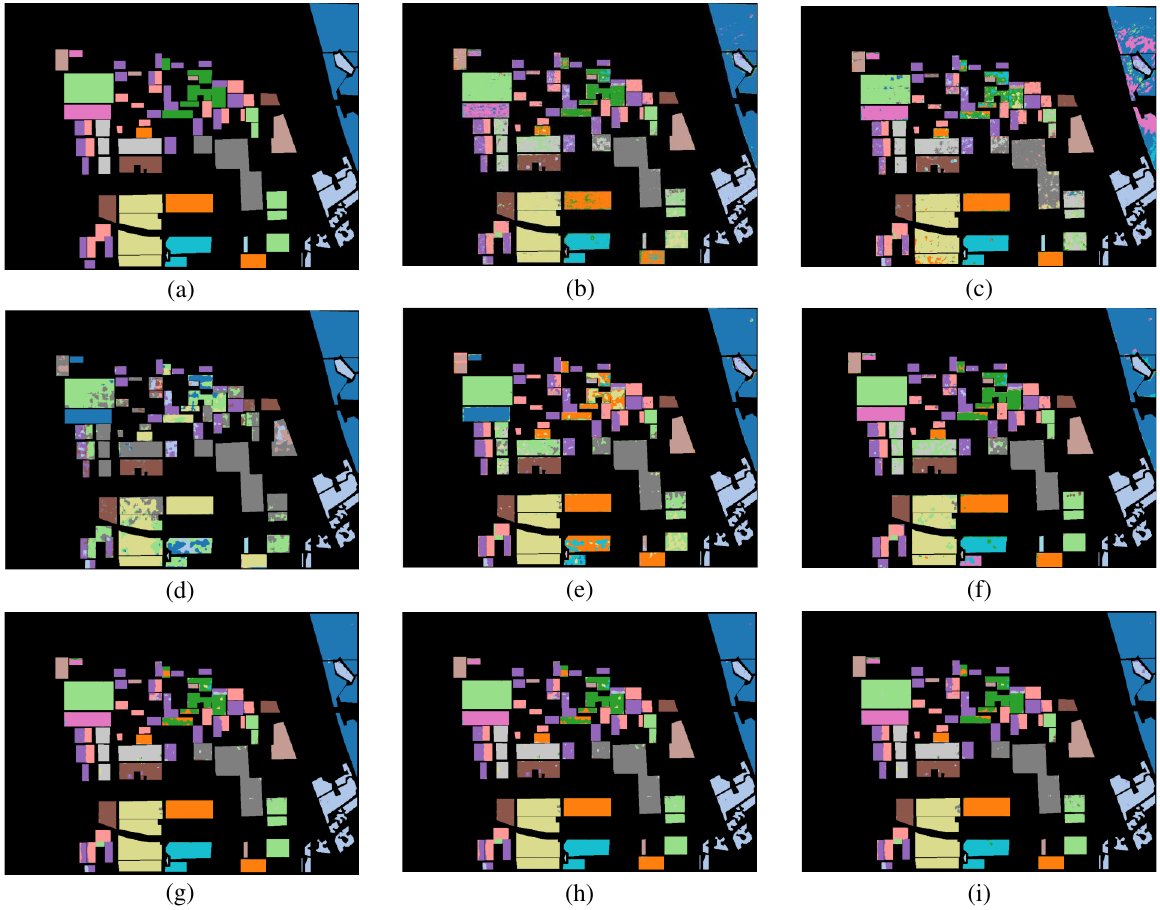}
\caption{Visualization of classification results for the  Flevoland dataset. (a) Ground Truth. (b) PCLNet. (c) CV-3DCNN. (d) MSFMamba. (e) PiCL. (f) PolSAR-Former. (g) CV-ASDF2Net. (h) SSL-MBC. (i) CV-SSMNet.}
\label{fig:8}
\end{figure*}

\begin{table*}[t]
\caption{Classification comparison results for the Flevoland dataset. The training set and test set represent the number of training samples and test samples for each class, respectively, and the samples were selected through a spatially disjoint partition. The best and second-best results are highlighted in bold and underlined, respectively.}
\label{tab:Table1}
\centering
\scriptsize
\renewcommand{\arraystretch}{1.10}
\setlength{\tabcolsep}{2.2pt}
\begin{tabular*}{\textwidth}{@{\extracolsep{\fill}}lrrcccccccc}
\toprule
Class & Train & Test & PCLNet & CV-3DCNN & MSFMamba & PiCL & PolSAR-Former & CV-ASDF2Net & SSL-MBC & CV-SSMNet \\
\midrule
Water      & 293 & 24499 & 96.67 & 98.69 & 96.69 & 99.15 & 99.01 & 98.82 & \underline{99.17} & \textbf{100.00} \\
Forest     & 159 & 12942 & 98.11 & \underline{98.87} & 95.78 & 97.48 & 96.29 & 98.33 & 98.61 & \textbf{100.00} \\
Lucerne    & 112 & 9940  & 91.76 & \underline{97.67} & 80.55 & 97.19 & 95.34 & 95.64 & \textbf{98.20} & 86.42 \\
Grass      & 103 & 6741  & 91.48 & 88.90 & 82.24 & \underline{96.06} & 88.87 & 94.84 & 95.18 & \textbf{100.00} \\
Rapeseed   & 219 & 19353 & 88.05 & \textbf{98.18} & 80.37 & 91.68 & \underline{97.04} & 95.13 & 95.13 & 95.06 \\
Beet       & 148 & 12259 & 94.58 & 89.14 & 85.90 & \underline{96.39} & 94.83 & 93.97 & \textbf{97.21} & 72.97 \\
Potatoes   & 214 & 17892 & 84.75 & 86.65 & 86.88 & \underline{95.14} & 91.58 & 91.95 & \textbf{98.25} & 83.23 \\
Peas       & 104 & 7999  & \textbf{99.98} & 96.97 & 91.08 & \underline{99.12} & 96.75 & 92.72 & 98.61 & 89.30 \\
Stem beans & 85  & 6718  & 92.03 & \underline{98.02} & 87.49 & 88.96 & 94.43 & 97.23 & 90.09 & \textbf{100.00} \\
Bare Soil  & 64  & 5496  & 88.90 & 93.91 & 89.64 & 92.99 & \underline{94.68} & 91.56 & 93.65 & \textbf{100.00} \\
Wheat      & 177 & 14759 & 97.73 & 88.07 & 76.56 & 95.85 & 95.15 & \underline{99.51} & 97.81 & \textbf{100.00} \\
Wheat 2    & 107 & 7268  & 92.19 & 95.29 & 92.15 & \underline{95.97} & 95.30 & 91.92 & 95.69 & \textbf{100.00} \\
Wheat 3    & 221 & 19287 & 90.64 & \underline{98.97} & 96.78 & 98.13 & 96.53 & 95.39 & 97.82 & \textbf{100.00} \\
Barley     & 74  & 6596  & 98.62 & 97.06 & 93.27 & 89.94 & \underline{98.36} & 92.23 & 97.65 & \textbf{100.00} \\
Buildings  & 6   & 416   & 99.09 & 79.41 & 87.65 & 98.79 & 81.83 & 97.21 & \underline{99.12} & \textbf{100.00} \\
\midrule
\multicolumn{3}{l}{OA (\%)} & $92.09\pm0.22$ & $94.72\pm1.32$ & $90.12\pm1.55$ & $95.10\pm0.18$ & $95.87\pm1.39$ & $95.33\pm0.72$ & $\underline{96.72\pm0.04}$ & $\mathbf{97.56\pm0.45}$ \\
\multicolumn{3}{l}{AA (\%)} & $93.63\pm0.15$ & $93.72\pm1.10$ & $88.20\pm1.42$ & $95.52\pm0.19$ & $94.40\pm1.43$ & $95.10\pm0.21$ & $\mathbf{96.81\pm0.06}$ & $\underline{95.13\pm0.23}$ \\
\multicolumn{3}{l}{Kappa$\times$100} & $91.37\pm0.26$ & $92.16\pm1.13$ & $89.56\pm1.77$ & $94.65\pm0.22$ & $95.49\pm1.45$ & $93.68\pm0.80$ & $\underline{96.42\pm0.05}$ & $\mathbf{97.05\pm0.36}$ \\
\bottomrule
\end{tabular*}
\end{table*}

As shown in \textcolor{blue}{Fig.~9}, the San Francisco results show clear differences among the compared methods in complex urban scenes. PCLNet, CV-3DCNN, and MSFMamba can identify major categories but still suffer from misclassifications and noise near building edges and transition areas. PiCL, PolSAR-Former, CV-ASDF2Net, and SSL-MBC improve spatial continuity to varying degrees, whereas CV-SSMNet produces maps closest to the ground truth, with sharper structural boundaries and fewer errors around buildings, roads, and small objects.

The quantitative results in \textcolor{blue}{Table~\ref{tab:Table2}} further confirm this advantage. CV-SSMNet achieves the highest OA of 97.02\% and Kappa$\times$100 of 95.06, outperforming CV-ASDF2Net by 1.22\% in OA and 1.04 in Kappa$\times$100. It also obtains the best accuracies for Urban and Vegetation. Although its AA is slightly lower than SSL-MBC due to weaker performance on Water and Mountain, CV-SSMNet shows stronger overall consistency and reliability.

\begin{figure*}[!t]
\centering
\captionsetup{justification=raggedright,singlelinecheck=false}
\includegraphics[width=\textwidth]{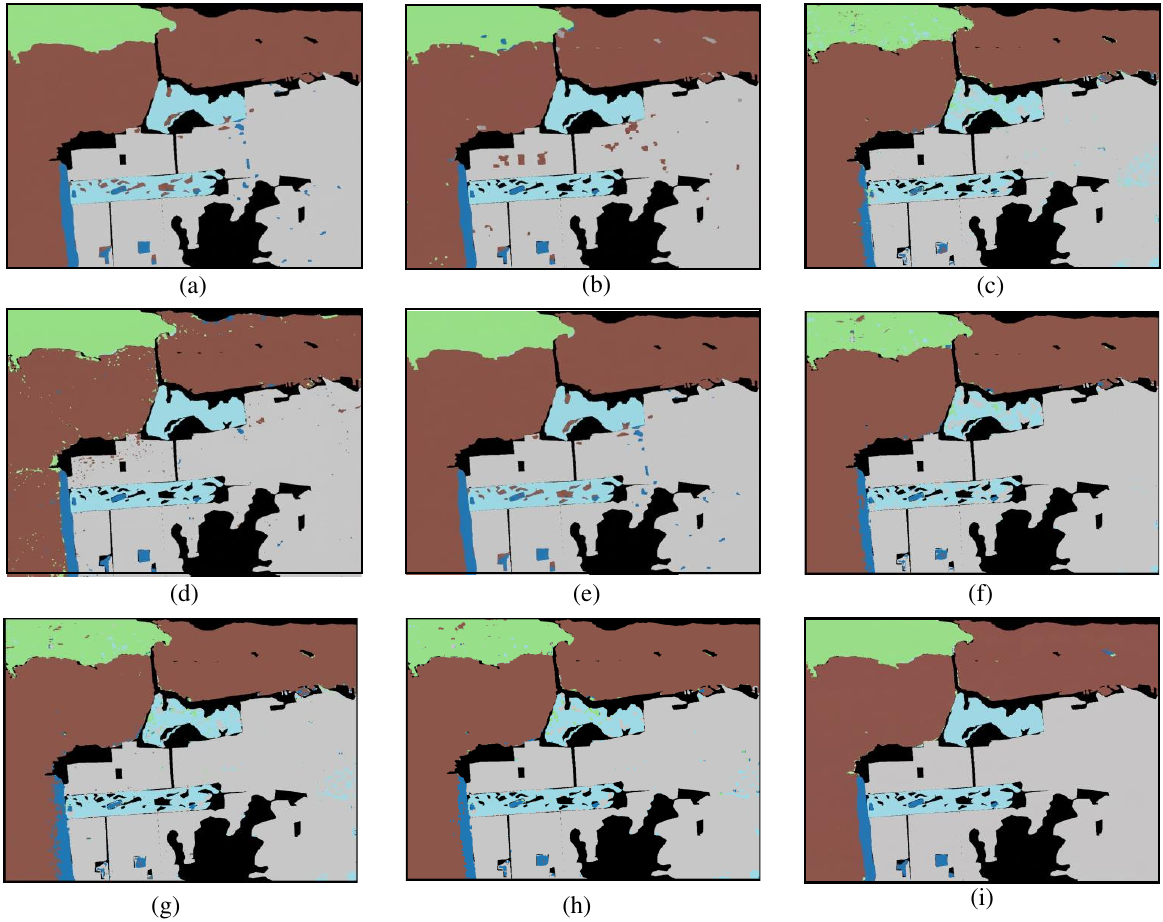}
\caption{Visualization of classification results for the San Francisco dataset. (a) Ground Truth. (b) PCLNet. (c) CV-3DCNN. (d) MSFMamba. (e) PiCL. (f) PolSAR-Former. (g) CV-ASDF2Net. (h) SSL-MBC. (i) CV-SSMNet.}
\label{fig:9}
\end{figure*}

\begin{table*}[t]
\caption{Classification comparison results for the San Francisco dataset. The training set and test set represent the number of training samples and test samples for each class, respectively, and the samples were selected through a spatially disjoint partition. The best and second-best results are highlighted in bold and underlined, respectively.}
\label{tab:Table2}
\centering
\scriptsize
\renewcommand{\arraystretch}{1.10}
\setlength{\tabcolsep}{2.2pt}
\begin{tabular*}{\textwidth}{@{\extracolsep{\fill}}lrrcccccccc}
\toprule
Class & Train & Test & PCLNet & CV-3DCNN & MSFMamba & PiCL & PolSAR-Former & CV-ASDF2Net & SSL-MBC & CV-SSMNet \\
\midrule
Water      & 3296 & 283087 & 87.50 & 86.02 & 64.53 & \underline{95.61} & 82.81 & 85.85 & \textbf{96.71} & 86.57 \\
Urban      & 3428 & 278242 & 92.49 & \underline{96.84} & 93.53 & 93.43 & 91.72 & 92.34 & 94.05 & \textbf{98.07} \\
Vegetation & 536  & 42961  & 93.10 & 97.62 & 98.97 & 95.94 & \underline{99.24} & 93.43 & 95.19 & \textbf{99.40} \\
Bare Soil  & 138  & 9528   & 89.65 & \textbf{98.24} & 97.11 & 92.31 & 95.93 & \underline{97.63} & 95.38 & 97.02 \\
Mountain   & 628  & 52882  & 80.70 & \textbf{87.45} & 76.40 & 85.91 & 69.57 & 76.06 & \underline{87.42} & 83.10 \\
\midrule
\multicolumn{3}{l}{OA (\%)} & $90.69\pm0.17$ & $93.23\pm1.76$ & $95.66\pm1.67$ & $93.52\pm0.09$ & $94.98\pm1.45$ & $\underline{95.80\pm0.12}$ & $94.09\pm0.10$ & $\mathbf{97.02\pm0.82}$ \\
\multicolumn{3}{l}{AA (\%)} & $88.69\pm4.30$ & $\underline{93.23\pm1.64}$ & $86.11\pm2.03$ & $92.64\pm0.21$ & $87.85\pm1.48$ & $89.06\pm0.03$ & $\mathbf{93.75\pm0.08}$ & $92.83\pm1.40$ \\
\multicolumn{3}{l}{Kappa$\times$100} & $86.03\pm0.32$ & $93.56\pm1.23$ & $93.17\pm1.48$ & $90.08\pm0.19$ & $92.12\pm1.49$ & $\underline{94.02\pm1.32}$ & $91.83\pm0.25$ & $\mathbf{95.06\pm1.05}$ \\
\bottomrule
\end{tabular*}
\end{table*}

As shown in \textcolor{blue}{Fig.~10}, the Oberpfaffenhofen visualization results reveal clear differences in regional consistency and boundary preservation. PCLNet, CV-3DCNN, and MSFMamba produce generally accurate maps but still contain speckle noise and local misclassifications, while PiCL, PolSAR-Former, CV-ASDF2Net, and SSL-MBC improve regional integrity to different extents. In contrast, CV-SSMNet aligns more closely with the ground truth, producing cleaner regions, sharper boundaries, and better structural continuity.

The quantitative results in \textcolor{blue}{Table~\ref{tab:Table3}} further support this observation. CV-SSMNet achieves the best OA of 96.07\% and Kappa$\times$100 of 94.72, with the highest accuracy for Build-Up Areas and competitive performance on Wood Land and Open Areas. Although its AA is slightly lower than CV-3DCNN, CV-SSMNet provides the most balanced performance across OA, AA, and kappa, indicating effective modeling of structural consistency and regional dependencies.

\begin{figure*}[!t]
\centering
\captionsetup{justification=raggedright,singlelinecheck=false}
\includegraphics[width=\textwidth]{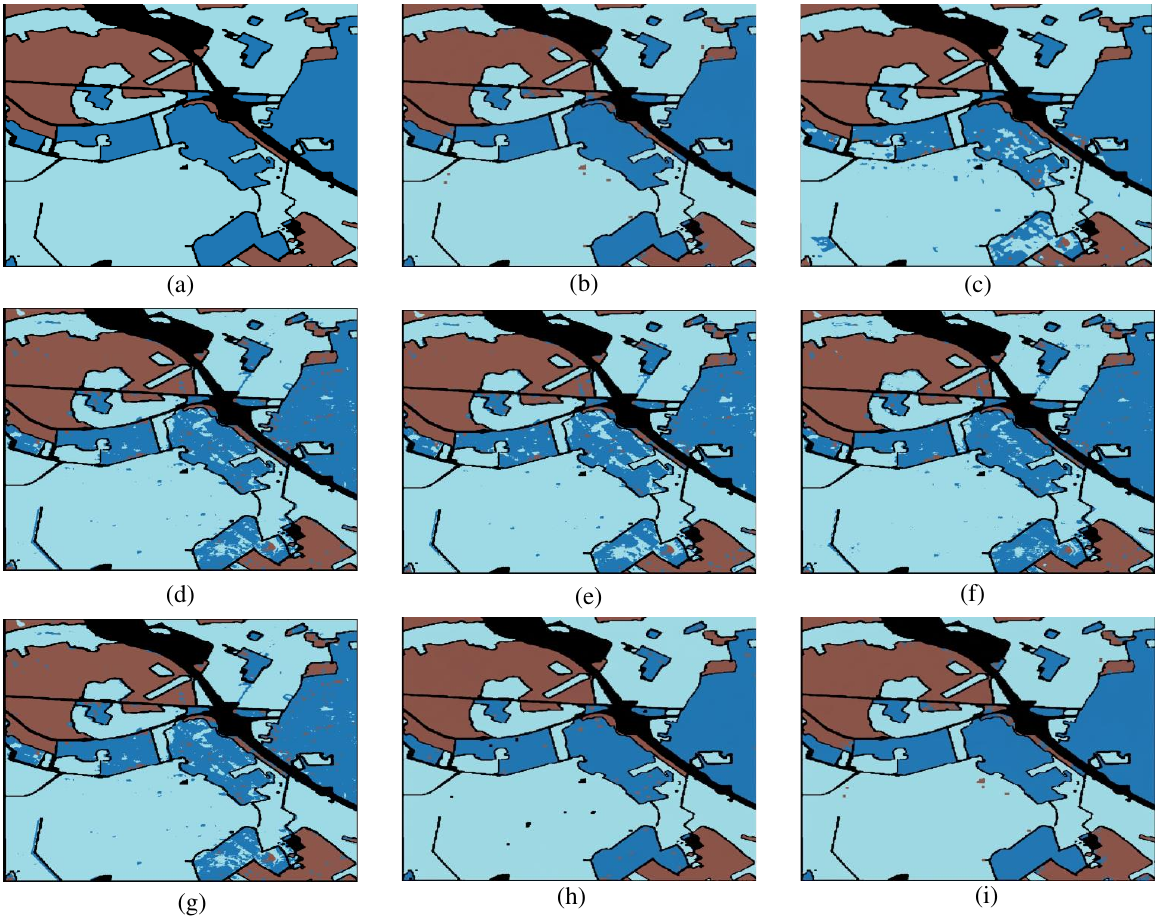}
\caption{Visualization of the classification results for the Oberpfaffenhofen dataset. (a) Ground Truth. (b) PCLNet. (c) CV-3DCNN. (d) MSFMamba. (e) PiCL. (f) PolSAR-Former. (g) CV-ASDF2Net. (h) SSL-MBC. (i) CV-SSMNet.}
\label{fig:10}
\end{figure*}

\begin{table*}[t]
\caption{Classification comparison results for the Oberpfaffenhofen dataset. The training set and test set represent the number of training samples and test samples for each class, respectively, and the samples were selected through a spatially disjoint partition. The best and second-best results are highlighted in bold and underlined, respectively.}
\label{tab:Table3}
\centering
\scriptsize
\renewcommand{\arraystretch}{1.10}
\setlength{\tabcolsep}{2.2pt}
\begin{tabular*}{\textwidth}{@{\extracolsep{\fill}}lrrcccccccc}
\toprule
Class & Train & Test & PCLNet & CV-3DCNN & MSFMamba & PiCL & PolSAR-Former & CV-ASDF2Net & SSL-MBC & CV-SSMNet \\
\midrule
Build-Up Areas & 3281 & 265658 & 82.53 & \underline{94.39} & 66.21 & 77.36 & 91.85 & 92.50 & 90.45 & \textbf{95.20} \\
Wood Land      & 2467 & 200557 & 86.65 & 95.46 & \textbf{99.63} & 89.08 & \underline{96.00} & 92.37 & 89.19 & 95.17 \\
Open Areas     & 7369 & 619531 & 89.41 & \underline{97.23} & \textbf{98.85} & 93.79 & 92.31 & 89.50 & 83.83 & 92.72 \\
\midrule
\multicolumn{3}{l}{OA (\%)} & $87.17\pm0.11$ & $95.69\pm1.47$ & $91.26\pm1.68$ & $88.80\pm0.37$ & $\underline{95.76\pm1.47}$ & $92.50\pm0.12$ & $90.45\pm0.10$ & $\mathbf{96.07\pm0.48}$ \\
\multicolumn{3}{l}{AA (\%)} & $86.20\pm0.47$ & $\underline{94.13\pm1.59}$ & $88.23\pm1.71$ & $86.75\pm0.50$ & $93.39\pm1.59$ & $91.46\pm0.15$ & $87.82\pm0.30$ & $\mathbf{94.36\pm0.30}$ \\
\multicolumn{3}{l}{Kappa$\times$100} & $78.47\pm0.35$ & $93.54\pm1.54$ & $85.14\pm1.87$ & $86.96\pm1.00$ & $\underline{93.76\pm1.54}$ & $89.50\pm0.18$ & $83.83\pm0.29$ & $\mathbf{94.72\pm0.22}$ \\
\bottomrule
\end{tabular*}
\end{table*}

\subsection{Effect of Input Representation: Complex-Valued PolSAR Input vs. 107-D Real-Valued Features}

To verify whether preserving the native complex-valued structure is beneficial, CV-SSMNet is compared with a real-valued baseline (RV-SSMNet) using 107-dimensional handcrafted polarimetric features~\cite{ref50} as input on the Flevoland dataset. As reported in \textcolor{blue}{Table~\ref{tab:input_representation}}, employing 107-dimensional handcrafted features, On the Flevoland dataset, CV-SSMNet outperforms the RV-SSMNet baseline,
suggesting that preserving the native complex-valued covariance representation
is beneficial in this setting. In contrast, direct complex-domain modeling allows CV-SSMNet to more realistically capture long-range dependencies and physically meaningful feature interactions. CV-SSMNet outperforms RV-SSMNet by 2.86, 2.90, and 2.84 percentage points in OA, AA, and kappa coefficients, respectively, indicating that preserving the native complex-valued representation of PolSAR data is more effective than converting it to high-dimensional real-valued descriptors.

\begin{table*}[t]
\caption{Comparison between complex-valued input and 107-D real-valued feature input on the Flevoland dataset.}
\label{tab:input_representation}
\centering
\scriptsize
\renewcommand{\arraystretch}{1.10}
\setlength{\tabcolsep}{4pt}
\begin{tabular*}{\textwidth}{@{\extracolsep{\fill}}lllccc}
\toprule
Method & Input type & Representation & OA (\%) & AA (\%) & Kappa$\times$100 \\
\midrule
RV-SSMNet & Real-valued & 107-D handcrafted polarimetric features
& $94.70\pm1.25$ & $92.23\pm2.03$ & $94.21\pm1.76$ \\
CV-SSMNet & Complex-valued & Complex-valued covariance representation
& $\mathbf{97.56\pm0.45}$ & $\mathbf{95.13\pm0.23}$ & $\mathbf{97.05\pm0.36}$ \\
\bottomrule
\end{tabular*}
\end{table*}

\subsection{Ablation Experiments}

\begin{table*}[t]
\caption{Ablation results of different components, prior settings, and scanning strategies on the three datasets. The best and second-best results within each ablation group are highlighted in bold and underlined, respectively.}
\label{tab:ablation_all}
\centering
\scriptsize
\setlength{\tabcolsep}{2.6pt}
\renewcommand{\arraystretch}{1.08}
\begin{tabular*}{\textwidth}{@{\extracolsep{\fill}}lccccccccc}
\toprule
\multirow{2}{*}{Method} & \multicolumn{3}{c}{Flevoland} & \multicolumn{3}{c}{San Francisco} & \multicolumn{3}{c}{Oberpfaffenhofen} \\
\cmidrule(lr){2-4}\cmidrule(lr){5-7}\cmidrule(lr){8-10}
& OA (\%) & AA (\%) & Kappa$\times$100 & OA (\%) & AA (\%) & Kappa$\times$100 & OA (\%) & AA (\%) & Kappa$\times$100 \\
\midrule
\multicolumn{10}{l}{\textit{A) Component ablation}} \\
Base & 95.33 & 89.90 & 93.68 & 95.80 & \underline{91.46} & 94.02 & 92.50 & 92.37 & 89.50 \\
Base + CV-SSM & 95.31 & 91.46 & 93.66 & 95.82 & 90.05 & \underline{94.82} & \underline{94.61} & \underline{93.61} & \underline{90.35} \\
Base + SFM & \underline{95.66} & \underline{94.20} & \underline{95.25} & \underline{96.26} & \textbf{91.85} & 94.23 & 94.23 & 93.22 & 89.69 \\
Base + CV-SSM + SFM & \textbf{97.56} & \textbf{95.13} & \textbf{97.05} & \textbf{97.02} & 92.83 & \textbf{95.06} & \textbf{96.07} & \textbf{94.36} & \textbf{94.72} \\
\midrule
\multicolumn{10}{l}{\textit{B) Prior ablation}} \\
Only $H/A/\alpha$ & 96.54 & 93.26 & \underline{96.31} & 96.78 & 91.17 & 94.64 & 94.68 & 94.76 & 93.58 \\
Only $P_s/P_d/P_v$ & \underline{97.11} & \underline{94.72} & 95.84 & \underline{96.92} & \textbf{91.59} & \textbf{95.60} & \underline{95.77} & \underline{95.42} & \textbf{95.06} \\
Only Span & 95.71 & 93.45 & 95.32 & 96.28 & 90.76 & 94.15 & 94.32 & 94.26 & 93.91 \\
All priors & \textbf{97.56} & \textbf{95.13} & \textbf{97.05} & \textbf{97.02} & \underline{92.83} & \underline{95.06} & \textbf{96.07} & \textbf{94.36} & \underline{94.72} \\
\midrule
\multicolumn{10}{l}{\textit{C) Scanning direction ablation}} \\
Horizontal scanning & 94.23 & 92.07 & 93.94 & 95.45 & 90.05 & 93.68 & \underline{95.62} & \underline{95.42} & 93.85 \\
Vertical scanning & \underline{95.14} & \underline{92.34} & \underline{94.15} & \underline{96.02} & \underline{90.33} & \underline{94.26} & 94.37 & 94.21 & \underline{93.89} \\
All directions & \textbf{97.56} & \textbf{95.13} & \textbf{97.05} & \textbf{97.02} & \textbf{92.83} & \textbf{95.06} & \textbf{96.07} & \textbf{94.36} & \textbf{94.72} \\
\bottomrule
\end{tabular*}
\end{table*}

\label{sec:ablation}

To dissect the contribution of each design choice in CV-SSMNet, a unified ablation study is conducted on three benchmark datasets. 
All variants share the same training protocol and spatial 
block-based partitioning to ensure a fair comparison. 
The aggregated results, including (A) component ablation, 
(B) prior ablation, and (C) scanning-direction ablation, 
are summarized in \textcolor{blue}{Table~\ref{tab:ablation_all}}.

\subsubsection{Effect of Different Modules}
CV-ASDF2Net is adopted as the baseline (\textit{Base}), and the proposed CV-SSM and SFM modules are progressively incorporated. As shown in \textcolor{blue}{Table~\ref{tab:ablation_all}-A}, the two modules provide complementary improvements. CV-SSM mainly enhances long-range contextual modeling, leading to clear gains on Oberpfaffenhofen, where OA and AA increase from 92.50\% and 92.37\% to 94.61\% and 93.61\%, respectively. SFM brings more consistent class-balanced improvements, especially on Flevoland, where AA increases from 89.90\% to 94.20\%, indicating that scattering priors are effective for fine-grained crop discrimination. When CV-SSM and SFM are combined, CV-SSMNet achieves the best OA, AA, and kappa$\times$100 among all component variants on the three datasets. Compared with the baseline, the OA improvements are 2.23\%, 1.22\%, and 3.57\% on Flevoland, San Francisco, and Oberpfaffenhofen, respectively, confirming the complementarity between complex-valued long-range modeling and scattering-aware feature modulation.

\subsubsection{Contribution of Each Prior Parameter}
\textcolor{blue}{Table~\ref{tab:ablation_all}-B} evaluates the influence of different physical prior groups. Using all priors achieves the highest OA on all three datasets, with 97.56\%, 97.02\%, and 96.07\% on Flevoland, San Francisco, and Oberpfaffenhofen, respectively. It also obtains the best AA on Flevoland and San Francisco, demonstrating that the joint use of $H/A/\alpha$, Freeman--Durden powers, and Span provides complementary scattering information. The Freeman--Durden powers $(P_s,P_d,P_v)$ show strong performance on San Francisco and Oberpfaffenhofen, particularly in kappa$\times$100 and AA, suggesting that surface, double-bounce, and volume scattering components are highly discriminative for structurally complex scenes. In contrast, using Span alone gives relatively lower performance, indicating that total backscattering power is insufficient without complementary polarimetric descriptors.

\subsubsection{Effect of Scanning Directions}
\textcolor{blue}{Table~\ref{tab:ablation_all}-C} compares horizontal, vertical, and bidirectional scanning strategies. Single-direction scanning exhibits dataset-dependent behavior: vertical scanning performs better on Flevoland and San Francisco, while horizontal scanning gives a higher AA on Oberpfaffenhofen. This indicates that unidirectional scanning is sensitive to the dominant spatial orientation of land-cover structures. By combining row-wise and column-wise dependencies, the bidirectional setting achieves the best OA and kappa$\times$100 on all three datasets, and the best AA on Flevoland and San Francisco. Although horizontal scanning obtains a slightly higher AA on Oberpfaffenhofen, the bidirectional strategy provides the most stable overall performance, demonstrating its ability to capture complementary spatial dependencies and reduce orientation bias in heterogeneous PolSAR scenes.

\subsubsection{Prior-injection strategy comparison.}
We further compare three prior-injection strategies: no prior, shallow
concatenation, and FiLM-based modulation. As shown in
\textcolor{blue}{Table~\ref{tab:physical_interpretability}}, FiLM modulation achieves the highest
intra-class scattering consistency (ISC) and inter-class discriminability (ISD).
Compared with shallow concatenation, FiLM modulation improves ISC by
approximately $2.9\times$ and ISD by approximately $2.4\times$. These results
suggest that conditional modulation is more effective than input-level fusion
for preserving physically meaningful scattering information during feature
evolution.

\textbf{Robustness under speckle noise.}
Finally, we evaluate the stability of different prior-injection strategies under
multiplicative speckle perturbation. The prediction consistency of FiLM
modulation remains $0.89$, compared with $0.74$ for shallow concatenation and
$0.67$ for the no-prior baseline. This indicates that scattering-aware FiLM
modulation improves robustness to noise-sensitive statistical variations and
helps maintain more stable predictions under degraded PolSAR observations.

\begin{table}[t]
\centering
\caption{Physical interpretability comparison of different prior-injection strategies.}
\label{tab:physical_interpretability}
\resizebox{\columnwidth}{!}{
\begin{tabular}{lcccc}
\hline
\textbf{Method} & \textbf{ISC}$\uparrow$ & \textbf{ISD}$\uparrow$ &
\textbf{SR}$\uparrow$ & \textbf{OA (\%)} \\
\hline
No Prior        & 0.23 & 1.45 & 0.67 & 95.33 \\
Shallow Concat  & 0.41 & 2.18 & 0.74 & 95.66 \\
FiLM Modulation & \textbf{0.68} & \textbf{3.52} & \textbf{0.89} & \textbf{97.56} \\
\hline
\end{tabular}
}
{\footnotesize Note: SR denotes speckle robustness.}
\end{table}

\subsection{Computational Complexity and Performance Stability}

\begin{table}[t]
\caption{Computational complexity and run-level performance stability of CV-SSMNet.}
\label{tab:complexity_significance}
\centering
\scriptsize
\renewcommand{\arraystretch}{1.08}
\setlength{\tabcolsep}{3pt}

\begin{tabular*}{\columnwidth}{@{\extracolsep{\fill}}lccccc}
\toprule
\multicolumn{6}{l}{\textit{A) Computational complexity comparison}} \\
\midrule
Method & Params & FLOPs & Mem. & Infer. & Training time \\
& (M) & (G) & (MB) & (ms) & (h) \\
\midrule
Base 
& \textbf{0.077} & \textbf{0.0100} & \textbf{512} & \textbf{0.40} & \textbf{8.5} \\
Base + CV-SSM 
& \underline{0.081} & \underline{0.0102} & \underline{640} & \underline{0.48} & \underline{10.2} \\
Base + SFM 
& 0.086 & 0.1050 & 704 & 0.52 & 10.8 \\
CV-SSMNet 
& 0.091 & 0.1120 & 712 & 0.55 & 11.5 \\
\bottomrule
\end{tabular*}

\vspace{1.0ex}

\begin{tabular*}{\columnwidth}{@{\extracolsep{\fill}}lccc}
\toprule
\multicolumn{4}{l}{\textit{B) Statistical significance test: Base vs. CV-SSMNet}} \\
\midrule
Dataset & $\chi^2$ & $p$-value & Significant \\
\midrule
Flevoland & 16.37 & 0.0005 & Yes \\
San Francisco & 34.86 & 0.00001 & Yes \\
Oberpfaffenhofen & 18.48 & 0.00002 & Yes \\
\bottomrule
\end{tabular*}
\end{table}
To further evaluate the efficiency and reliability of the proposed method, the computational cost of different model variants is reported, and a statistical significance test against the baseline is performed.
As shown in \textcolor{blue}{Table~\ref{tab:complexity_significance}-A}, reports the computational complexity of different model variants, while \textcolor{blue}{Table~\ref{tab:complexity_significance}-B} presents the statistical significance test between the baseline and CV-SSMNet.
CV-SSMNet introduces a small increase in parameters, while the FLOPs increase
due to the additional CV-SSM and modulation modules. Nevertheless, the absolute
computational cost remains low, with an inference time below 0.6 ms per sample.
Although FLOPs increase from 0.0100G to 0.1120G, the absolute computational cost remains low, with an inference time below 0.6 ms per sample.the overall computational overhead remains 
acceptable considering the consistent improvements in classification 
accuracy. These results indicate that the proposed long-range modeling 
and scattering-aware modulation modules improve representation capability 
without introducing prohibitive computational cost.

In addition, \textcolor{blue}{Table~\ref{tab:complexity_significance}-B} reports the 
statistical significance test between the baseline and CV-SSMNet on 
the three benchmark datasets. The resulting $p$-values are all below 
0.001, indicating that the performance gains achieved by CV-SSMNet are 
stable across five independent runs.

\subsection{P-band BIOMASS Application}

Beyond the three L-band benchmarks, CV-SSMNet is further evaluated on the BIOMASS Level-1b dataset as a large-scale P-band experiment. Compared with the L-band scenes, BIOMASS covers a much larger area and contains five land-cover categories, leading to stronger spatial heterogeneity and intra-class scattering variability. Since training and testing are performed within each dataset, this experiment is not a strict L-to-P cross-frequency transfer evaluation. Instead, it assesses the applicability of CV-SSMNet under different frequency-band and scene-scale conditions, especially in the presence of frequency-dependent scattering variations.

As shown in \textcolor{blue}{Table~\ref{tab:cross_band_results}}, CV-SSMNet achieves OA values of 97.56\%, 97.02\%, and 96.07\% on Flevoland, San Francisco, and Oberpfaffenhofen, respectively. On the P-band BIOMASS dataset, it obtains an OA of 88.87\%, an AA of 82.32\%, and a kappa$\times$100 of 84.57. Compared with the Base model, CV-SSMNet improves OA, AA, and kappa$\times$100 by 2.57, 2.86, and 0.99 points, respectively, indicating that CV-SSM and scattering-aware feature modulation remain effective in the large-scale P-band scenario.

The relatively lower accuracy on BIOMASS can be mainly attributed to three factors:
\begin{enumerate}
    \item P-band SAR has greater penetration depth than L-band SAR, resulting in different scattering responses and polarimetric statistics.
    \item The BIOMASS scene covers a larger geographical area, causing stronger spatial non-stationarity and intra-class variation.
    \item The spatial block-based partition reduces spatial leakage but increases the difficulty of cross-region classification.
\end{enumerate}

Despite these challenges, CV-SSMNet adapts to P-band observations through shared complex-valued representation learning and prior-conditioned modulation. The coherency matrix $\mathbf{T}_3$ provides a unified input form, dataset-specific channel normalization reduces band-dependent magnitude shifts, and SFM recalibrates intermediate features using the scattering priors $H$, $A$, $\alpha$, $P_s$, $P_d$, $P_v$, and $\mathrm{Span}$. The ablation results in \textcolor{blue}{Fig.~11} further demonstrate the effectiveness of CV-SSM and SFM on the large-scale BIOMASS scene.

\begin{table*}[t]
\caption{Multi-dataset evaluation on L-band benchmarks and the P-band BIOMASS dataset.}
\label{tab:cross_band_results}
\centering
\small
\setlength{\tabcolsep}{5pt}
\renewcommand{\arraystretch}{1.10}
\begin{tabular*}{\textwidth}{@{\extracolsep{\fill}}llcccccc}
\toprule
\multirow{2}{*}{Dataset} & \multirow{2}{*}{Model} & \multicolumn{3}{c}{Data description} & \multicolumn{3}{c}{Performance} \\
\cmidrule(lr){3-5}\cmidrule(lr){6-8}
& & Band & Classes & Image size & OA (\%) & AA (\%) & Kappa$\times$100 \\
\midrule
Flevoland        & CV-SSMNet & L & 15 & $750 \times 1024$   & 97.56 & 95.13 & 97.05 \\
San Francisco    & CV-SSMNet & L & 5  & $900 \times 1024$   & 97.02 & 92.83 & 95.06 \\
Oberpfaffenhofen & CV-SSMNet & L & 3  & $1300 \times 1200$  & 96.07 & 94.36 & 94.72 \\
BIOMASS          & CV-SSMNet & P & 5  & $21172 \times 1509$ & \textbf{88.87} & \textbf{82.32} & \textbf{84.57} \\
BIOMASS          & Base      & P & 5  & $21172 \times 1509$ & \underline{86.30} & \underline{79.46} & \underline{83.58} \\
\bottomrule
\end{tabular*}
\end{table*}

\begin{figure*}[!t]
\centering
\captionsetup{justification=raggedright,singlelinecheck=false}
\includegraphics[width=\textwidth]{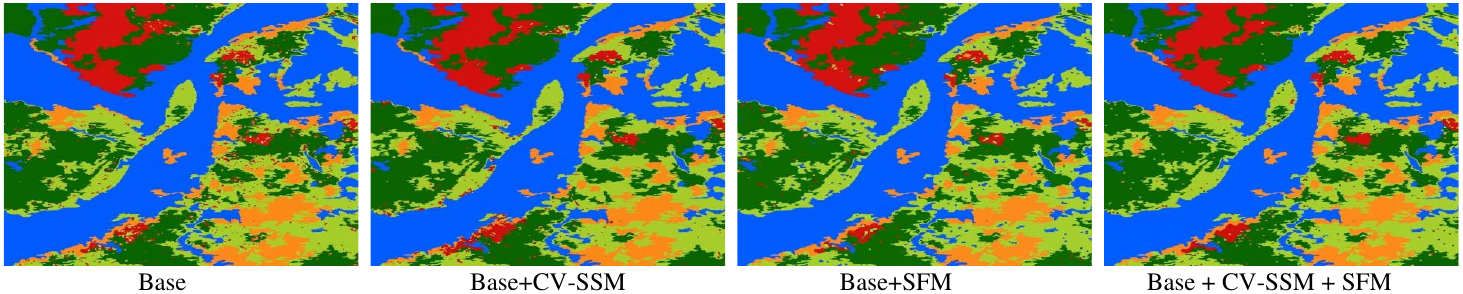}
\caption{Visualization of ablation classification results on the BIOMASS dataset. (a) Base. (b) Base + CV-SSM. (c) Base + SFM. (d) CV-SSMNet. The progressive incorporation of CV-SSM and SFM improves regional consistency, boundary preservation, and misclassification suppression.}
\label{fig:11}
\end{figure*}

\subsection{Feature Visualization Analysis}
\subsubsection{Feature visualization}

To evaluate the discriminative power of the learned features, t-SNE is employed to visualize the learned feature distributions. Specifically, in \textcolor{blue}{Fig.~12 (a)-(c)} , the features extracted by the Base model are scattered across the three datasets and exhibit substantial inter-class overlap. The inter-class mixing is particularly prominent in the areas marked by red ellipses. This indicates that the features extracted by the Base model have limited discriminative power, making it difficult to form clear and stable intra-class clustering and inter-class separation in the high-dimensional space, especially for land cover categories with similar scattering characteristics or complex structures, which easily leads to confusion. This insufficient feature representation also explains its limited performance in cross-regional or complex scenarios. In contrast, the CV-SSMNet feature visualization results shown in \textcolor{blue}{Fig.~12 (d)-(f)} demonstrate significant improvement. On all three datasets, samples of the same class exhibit a more compact clustering structure in the low-dimensional embedded space, and the separation between different classes is significantly increased. The areas that were severely overlapping in the Base model are effectively separated. This phenomenon indicates that the features learned by CV-SSMNet have stronger discriminative power and consistency.

\begin{figure}
	\centering
	\includegraphics[width=.9\columnwidth]{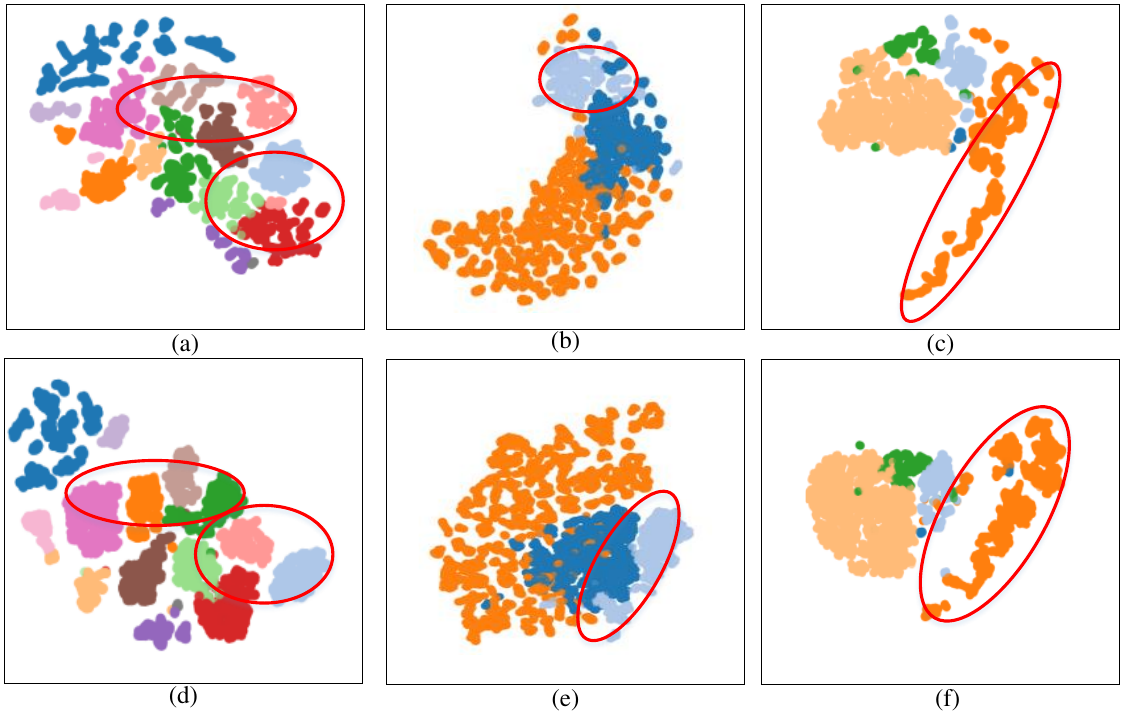}
	\caption{t-SNE feature visualization on three benchmark datasets. The first, second, and third columns correspond to Flevoland, Oberpfaffenhofen, and San Francisco, respectively. The top row shows the Base model, and the bottom row shows CV-SSMNet. CV-SSMNet yields more compact and separable feature distributions across different scenes.}
	\label{FIG:12}
\end{figure}

\subsubsection{Physics-Aware Activation and Layer-wise Feature Analysis}

To analyze how scattering priors and long-range modeling influence the learned representation, \textcolor{blue}{Fig.~13} visualizes the activation responses of the seven physical prior channels and their fused output. Different priors show spatial patterns broadly consistent with PolSAR scattering mechanisms. The $H$ channel responds strongly to regions with high scattering randomness, while $P_s$, $P_d$, and $P_v$ highlight surface-like, double-bounce, and volume-scattering areas, respectively. In contrast, $A$, $\alpha$, and $\mathrm{Span}$ provide smoother complementary contextual responses. The fused activation map integrates these scattering-dependent cues, indicating that SFM uses physical priors as conditional modulation signals rather than simple concatenated features.

\begin{figure}
	\centering
	\includegraphics[width=.9\columnwidth]{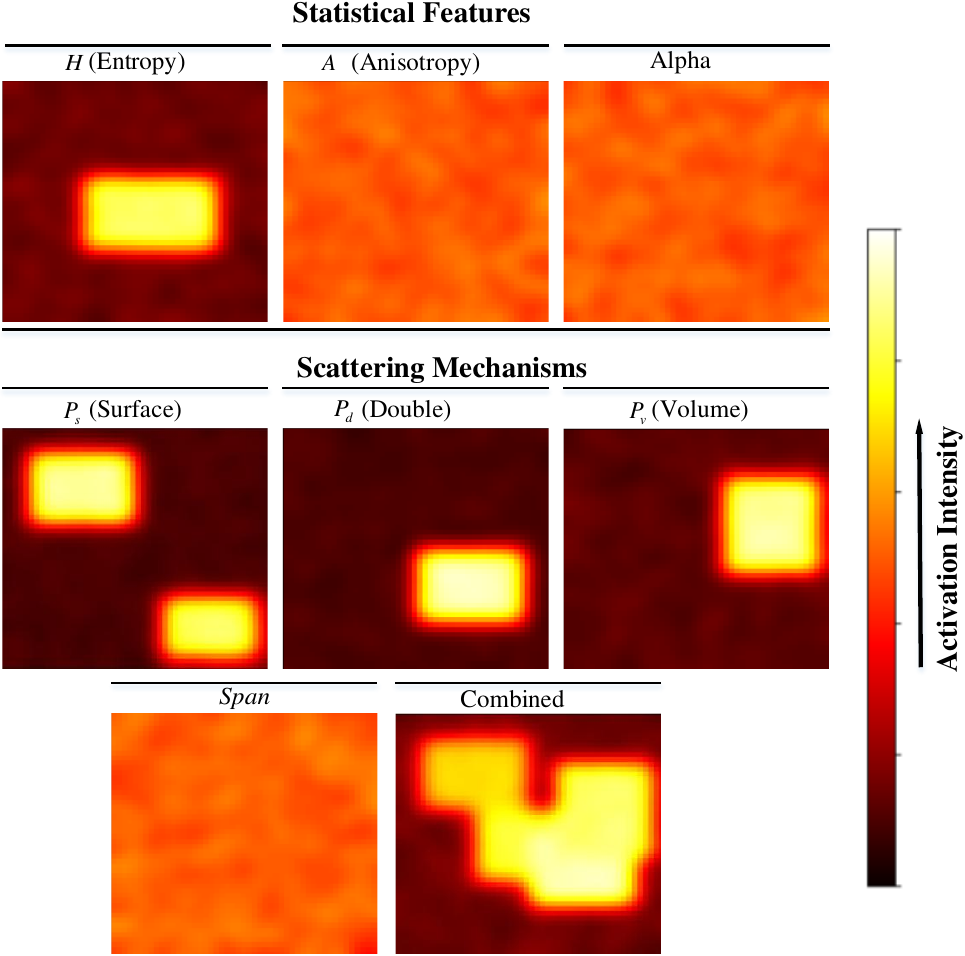}
	\caption{Prior-conditioned activation responses of the seven polarimetric scattering descriptors. The responses are broadly consistent with typical PolSAR scattering mechanisms, indicating that SFM learns physically meaningful modulation patterns rather than purely data-driven attention weights.}
	\label{FIG:13}
\end{figure}

\textcolor{blue}{Fig.~14} further compares intermediate feature maps of the baseline and CV-SSMNet across representative stages, including shallow convolution, middle convolution, deep convolution with CV-SSM, and final logits. Both models capture basic local structures at shallow layers, but CV-SSMNet produces stronger boundary responses and more uniform activations over heterogeneous regions. At deeper layers, the baseline tends to focus on central areas with weaker peripheral responses, whereas CV-SSMNet shows broader spatial coverage, confirming that complex-valued state-space modeling captures long-range dependencies more effectively. The output maps also exhibit sharper boundaries and more separable class responses, demonstrating improved regional consistency and boundary discriminability over the CV-ASDF2Net baseline.

\begin{figure}
	\centering
	\includegraphics[width=.9\columnwidth]{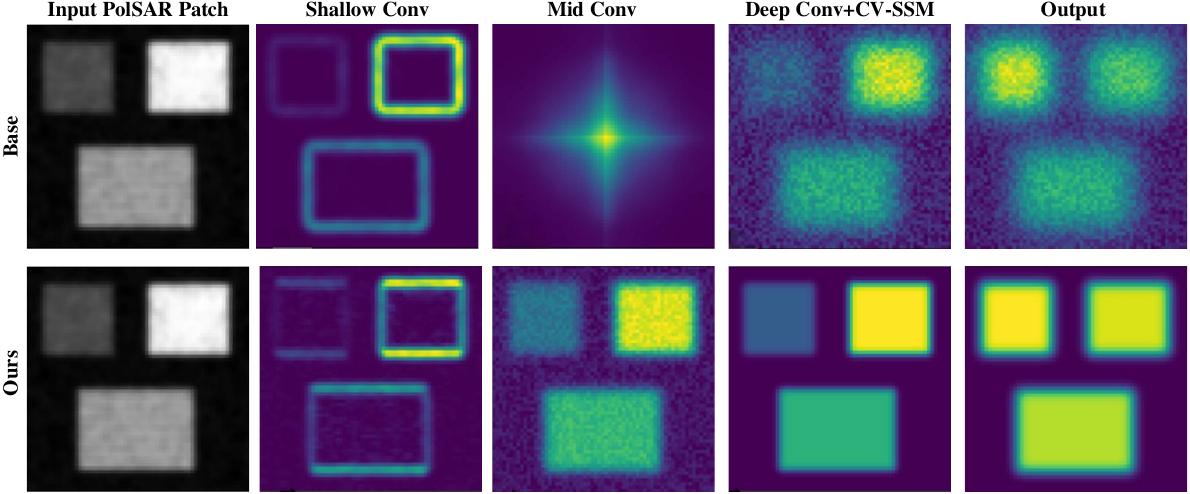}
	\caption{Feature maps at different layers of the baseline (top) and the proposed CV-SSMNet (bottom). Columns: input patch, shallow Conv, mid Conv, deep Conv + CV-SSM, and output.}
	\label{FIG:14}
\end{figure}

\section{Discussion and Conclusion}

This paper presents CV-SSMNet, a physics-aware complex-valued state-space
network for PolSAR image classification. Unlike methods that use polarimetric
descriptors only as shallow auxiliary inputs, CV-SSMNet encodes
$\{H,A,\alpha,P_s, \\ P_d,P_v,\mathrm{Span}\}$ as FiLM-style modulation signals to
guide the evolution of complex-valued features. By integrating complex-valued
convolution, CV-SSM-based long-range dependency modeling, and scattering-aware
feature modulation, the proposed framework learns physically guided
representations from local scattering structures to global spatial context.

The proposed physics-aware mechanism follows a soft-constraint strategy rather
than a hard physics-constrained formulation. The seven polarimetric priors are
not directly imposed through first-principles electromagnetic equations, such as
Maxwell equations or radiative transfer models. Instead, they serve as
physically meaningful conditioning variables for data-driven feature evolution.
This design preserves the flexibility of deep learning while introducing
scattering-mechanism awareness into intermediate feature modulation, making the
framework adaptable to heterogeneous land-cover scenes and extensible to other
polarimetric descriptors.

Experiments on three L-band benchmark datasets and an additional P-band BIOMASS
dataset demonstrate that CV-SSMNet achieves competitive OA, AA, and kappa
values. Ablation results show that CV-SSM and SFM provide complementary gains:
CV-SSM improves long-range contextual modeling, while SFM enhances
scattering-aware feature recalibration. Visualization results further indicate
that the learned prior-conditioned responses are broadly consistent with typical
PolSAR scattering mechanisms, supporting the physical interpretability of the
proposed modulation strategy.

Several limitations remain. The lower performance on BIOMASS suggests that
frequency-dependent scattering variations are not fully modeled, especially
wavelength-dependent penetration depth, volume scattering, and dielectric
response differences between L-band and P-band observations. In addition, the
fixed $13\times13$ patch size may limit scale adaptability, and the CV-SSM and
SFM modules introduce extra computational cost, although the absolute inference
time remains low. Decomposition-based priors may also be affected by speckle
noise, low-SNR observations, and decomposition errors.

Future work will explore band-aware prior encoding, adaptive multi-scale patch
sampling, efficient CV-SSM design, and uncertainty-aware prior gating. Another
direction is to incorporate explicit electromagnetic scattering constraints or
differentiable scattering models, moving from soft physics-guided modulation
toward stronger physics-constrained GeoAI representation learning. The code is
available at \url{https://github.com/Lucia-210/CV-SSMNet}.
\section*{Acknowledgments}
The authors gratefully acknowledge the providers of the datasets and the developers of the PolSARpro software. This study was conducted within the framework of the Dragon-6 Cooperation Programme between the European Space Agency (ESA) and the Ministry of Science and Technology (MOST) of China.

\bibliographystyle{cas-model2-names}
\bibliography{cas-refs}

@ARTICLE{ref1,
  author  = {Z. Wang and L. Zhao and Y. Wang and others},
  title   = {AIR-PolSAR-Seg-2.0: Polarimetric SAR ground terrain classification dataset for large-scale complex scenes},
  journal = {Journal of Radars},
  volume  = {14},
  number  = {2},
  pages   = {353--365},
  year    = {2025},
  doi     = {10.12000/JR24237}
}

@ARTICLE{ref2,
  author  = {S. Wang and Z. Sun and T. Bian and Y. Guo and L. Dai and Y. Guo and L. Jiao},
  title   = {CDFNet: Cross-domain feature fusion network for PolSAR terrain classification},
  journal = {IEEE Trans. Geosci. Remote Sens.},
  volume  = {63},
  pages   = {1--15},
  year    = {2025}
}

@ARTICLE{ref3,
  author  = {W. An and M. Lin},
  title   = {Generalized polarimetric entropy: Polarimetric information quantitative analyses of model-based incoherent polarimetric decomposition},
  journal = {IEEE Trans. Geosci. Remote Sens.},
  volume  = {59},
  number  = {3},
  pages   = {2041--2057},
  year    = {2021}
}

@ARTICLE{ref4,
  author  = {D. Zhuang and L. Zhang and B. Zou},
  title   = {Model-based polarimetric SAR target decomposition: A scheme to introduce repeat-pass PolInSAR coherence},
  journal = {IEEE Trans. Geosci. Remote Sens.},
  volume  = {62},
  pages   = {1--16},
  year    = {2024}
}

@ARTICLE{ref5,
  author  = {Q. Wu and B. Hou and Z. Wen and L. Jiao},
  title   = {Variational learning of mixture Wishart model for PolSAR image classification},
  journal = {IEEE Trans. Geosci. Remote Sens.},
  volume  = {57},
  number  = {1},
  pages   = {141--154},
  year    = {2019}
}

@ARTICLE{ref6,
  author  = {Y. Zhang and W. Wang and Z. Guo and N. Li},
  title   = {Enhanced PGA for dual-polarized ISAR imaging by exploiting Cloude--Pottier decomposition},
  journal = {IEEE Geosci. Remote Sens. Lett.},
  volume  = {21},
  pages   = {1--5},
  year    = {2024}
}

@ARTICLE{ref7,
  author  = {A. Freeman and S. L. Durden},
  title   = {A three-component scattering model for polarimetric SAR data},
  journal = {IEEE Trans. Geosci. Remote Sens.},
  volume  = {36},
  number  = {3},
  pages   = {963--973},
  year    = {1998}
}

@ARTICLE{ref8,
  author  = {X. Nie and H. Qiao and B. Zhang},
  title   = {A variational model for PolSAR data speckle reduction based on the Wishart distribution},
  journal = {IEEE Trans. Image Process.},
  volume  = {24},
  number  = {4},
  pages   = {1209--1222},
  year    = {2015}
}

@ARTICLE{ref9,
  author  = {J. S. Lee and others},
  title   = {Unsupervised classification using polarimetric decomposition and the complex Wishart classifier},
  journal = {IEEE Trans. Geosci. Remote Sens.},
  volume  = {37},
  number  = {5},
  pages   = {2249--2258},
  year    = {1999}
}

@ARTICLE{ref10,
  author  = {D. Xiao and C. Liu},
  title   = {PolSAR terrain classification based on fine-tuned dilated group-cross convolution neural network},
  journal = {Journal of Radars},
  volume  = {8},
  number  = {4},
  pages   = {479--489},
  year    = {2019},
  doi     = {10.12000/JR19039}
}

@ARTICLE{ref11,
  author  = {T. Hu and W. Li and X. Qin and others},
  title   = {Terrain classification of polarimetric synthetic aperture radar images based on deep learning and conditional random field model},
  journal = {Journal of Radars},
  volume  = {8},
  number  = {4},
  pages   = {471--478},
  year    = {2019},
  doi     = {10.12000/JR18065}
}

@ARTICLE{ref12,
  author  = {Z. Zhang and H. Wang and F. Xu and Y.-Q. Jin},
  title   = {Complex-valued convolutional neural network and its application in polarimetric SAR image classification},
  journal = {IEEE Trans. Geosci. Remote Sens.},
  volume  = {55},
  number  = {12},
  pages   = {7177--7188},
  year    = {2017}
}

@ARTICLE{ref13,
  author  = {Z. Kuang and H. Bi and F. Li and others},
  title   = {Polarimetry-inspired contrastive learning for class-imbalanced PolSAR image classification},
  journal = {IEEE Trans. Geosci. Remote Sens.},
  volume  = {62},
  pages   = {1--19},
  year    = {2024}
}

@ARTICLE{ref14,
  author  = {W. Li and H. Xia and B. Xi and Y. Wang and Y. He and Y. Han},
  title   = {SSL-MBC: Self-supervised learning with multibranch consistency for few-shot PolSAR image classification},
  journal = {IEEE J. Sel. Topics Appl. Earth Observ. Remote Sens.},
  volume  = {18},
  pages   = {4696--4710},
  year    = {2025},
  doi     = {10.1109/JSTARS.2025.3528529}
}

@ARTICLE{ref15,
  author  = {A. Jamali and S. K. Roy and A. Bhattacharya and P. Ghamisi},
  title   = {Local window attention transformer for polarimetric SAR image classification},
  journal = {IEEE Geosci. Remote Sens. Lett.},
  volume  = {20},
  pages   = {1--5},
  year    = {2023}
}

@ARTICLE{ref16,
  author  = {J. Ni and F. Zhang and Q. Yin and Y. Zhou and H.-C. Li and W. Hong},
  title   = {Random neighbor pixel-block-based deep recurrent learning for polarimetric SAR image classification},
  journal = {IEEE Trans. Geosci. Remote Sens.},
  volume  = {59},
  number  = {9},
  pages   = {7557--7569},
  year    = {2021}
}

@ARTICLE{ref17,
  author  = {J. Cheng and D. Xiang and Q. Yin and F. Zhang},
  title   = {A novel crop classification method based on the tensor-GCN for time-series PolSAR data},
  journal = {IEEE Trans. Geosci. Remote Sens.},
  volume  = {60},
  pages   = {1--14},
  year    = {2022},
  note    = {Art. no. 5237614}
}

@ARTICLE{ref18,
  author  = {Q. Yin and Z. Lin and W. Hu and C. L{\'o}pez-Mart{\'i}nez and J. Ni and F. Zhang},
  title   = {Crop classification of multitemporal PolSAR based on 3-D attention module with ViT},
  journal = {IEEE Geosci. Remote Sens. Lett.},
  volume  = {20},
  pages   = {1--5},
  year    = {2023}
}

@ARTICLE{ref19,
  author  = {M. Q. Alkhatib and M. S. Zitouni and M. Al-Saad and others},
  title   = {PolSAR image classification using shallow to deep feature fusion network with complex-valued attention},
  journal = {Scientific Reports},
  volume  = {15},
  number  = {1},
  pages   = {24315},
  year    = {2025}
}

@ARTICLE{ref20,
  author  = {Z. Kuang and K. Liu and H. Bi and F. Li},
  title   = {PolSAR image classification with complex-valued diffusion model as representation learners},
  journal = {IEEE Trans. Aerosp. Electron. Syst.},
  volume  = {61},
  pages   = {1--21},
  year    = {2025},
  doi     = {10.1109/TAES.2025.3572877}
}

@ARTICLE{ref21,
  author  = {M. Q. Alkhatib},
  title   = {PolSAR image classification using a hybrid complex-valued network (HybridCVNet)},
  journal = {IEEE Geosci. Remote Sens. Lett.},
  volume  = {21},
  pages   = {1--5},
  year    = {2024}
}

@ARTICLE{ref22,
  author  = {M. Liu and L. Jiao and X. Liu and L. Li and F. Liu and S. Yang and Y. Guo and P. Chen},
  title   = {C2N2: Complex-valued contourlet neural network},
  journal = {IEEE J. Sel. Topics Appl. Earth Observ. Remote Sens.},
  volume  = {17},
  pages   = {4478--4491},
  year    = {2024}
}

@ARTICLE{ref23,
  author  = {M. Imani},
  title   = {Low frequency and radar's physical based features for improvement of convolutional neural networks for PolSAR image classification},
  journal = {Egypt. J. Remote Sens. Space Sci.},
  volume  = {25},
  number  = {1},
  pages   = {55--62},
  year    = {2022},
  doi     = {10.1016/j.ejrs.2021.12.007}
}

@ARTICLE{ref24,
  author  = {A. H. Ghazvinizadeh and M. Imani and H. Ghassemian},
  title   = {Residual network based on entropy-anisotropy-alpha target decomposition for polarimetric SAR image classification},
  journal = {Earth Sci. Informat.},
  volume  = {16},
  number  = {1},
  pages   = {357--366},
  year    = {2023},
  doi     = {10.1007/s12145-023-00944-6}
}

@ARTICLE{ref25,
  author  = {J. Shi and M. Nie and S. Ji and C. Shi and H. Liu and H. Jin},
  title   = {Polarimetric synthetic aperture radar image classification based on double-channel convolution network and edge-preserving Markov random field},
  journal = {Remote Sens.},
  volume  = {15},
  number  = {23},
  year    = {2023},
  note    = {Art. no. 5458},
  doi     = {10.3390/rs15235458}
}

@ARTICLE{ref26,
  author  = {S. Zhang and L. Cui and Z. Dong and W. An},
  title   = {A deep learning classification scheme for PolSAR image based on polarimetric features},
  journal = {Remote Sens.},
  volume  = {16},
  number  = {10},
  year    = {2024},
  note    = {Art. no. 1676},
  doi     = {10.3390/rs16101676}
}

@ARTICLE{ref27,
  author  = {Z. Kuang and H. Bi and F. Li and C. Xu},
  title   = {ECP-Mamba: An efficient multiscale self-supervised contrastive learning method with state space model for PolSAR image classification},
  journal = {IEEE Trans. Geosci. Remote Sens.},
  volume  = {63},
  pages   = {1--18},
  year    = {2025}
}

@ARTICLE{ref28,
  author  = {A. Gu and K. Goel and C. R{\'e}},
  title   = {Efficiently modeling long sequences with structured state spaces},
  journal = {arXiv preprint arXiv:2111.00396},
  year    = {2021}
}

@ARTICLE{ref29,
  author  = {A. Gu and T. Dao},
  title   = {Mamba: Linear-time sequence modeling with selective state spaces},
  journal = {arXiv preprint arXiv:2312.00752},
  year    = {2023}
}

@ARTICLE{ref31,
  author  = {L. Zhu and B. Liao and Q. Zhang and others},
  title   = {Vision Mamba: Efficient visual representation learning with bidirectional state space model},
  journal = {arXiv preprint arXiv:2401.09417},
  year    = {2024}
}

@INPROCEEDINGS{ref32,
  author    = {Y. Liu and Y. Tian and Y. Zhao and others},
  title     = {VMamba: Visual state space model},
  booktitle = {Adv. Neural Inf. Process. Syst.},
  volume    = {37},
  pages     = {103031--103063},
  year      = {2024}
}

@ARTICLE{ref33,
  author  = {F. Gao and X. Jin and X. Zhou and J. Dong and Q. Du},
  title   = {MSFMamba: Multi-scale feature fusion state space model for multi-source remote sensing image classification},
  journal = {IEEE Trans. Geosci. Remote Sens.},
  year    = {2025},
  note    = {Early Access}
}

@ARTICLE{ref34,
  author  = {W. Han and H. Fu and J. Zhu and S. Zhang and Q. Xie and J. Hu},
  title   = {A polarimetric projection-based scattering characteristics extraction tool and its application to PolSAR image classification},
  journal = {ISPRS J. Photogramm. Remote Sens.},
  volume  = {202},
  pages   = {314--333},
  year    = {2023}
}

@ARTICLE{ref35,
  author  = {C. Hu and Y. Wang and X. Sun and S. Quan and D. Xiang},
  title   = {Model-Based Polarimetric Target Decomposition With Power Redistribution for Urban Areas},
  journal = {IEEE J. Sel. Topics Appl. Earth Observ. Remote Sens.},
  volume  = {16},
  pages   = {8795--8808},
  year    = {2023}
}

@ARTICLE{ref36,
  author  = {D. Duan and Y. Wang and Y. Zhang},
  title   = {The critical role of cross-polarized backscatter in understanding L-band PolSAR data in forested and urban environments},
  journal = {Remote Sens. Environ.},
  volume  = {311},
  year    = {2024},
  note    = {Art. no. 114265}
}

@ARTICLE{ref37,
  author  = {Z. Guo and H. Zhang and J. Ge and Z. Shi and L. Xu and Y. Tang and F. Wu and Y. Wang and C. Wang},
  title   = {Built-up area extraction in PolSAR imagery using real-complex polarimetric features and feature fusion classification network},
  journal = {Int. J. Appl. Earth Obs. Geoinf.},
  volume  = {134},
  year    = {2024},
  note    = {Art. no. 104144}
}

@ARTICLE{ref38,
  author  = {M. Imani},
  title   = {Attention based network for fusion of polarimetric and contextual features for polarimetric synthetic aperture radar image classification},
  journal = {Eng. Appl. Artif. Intell.},
  volume  = {139},
  year    = {2025},
  note    = {Art. no. 109665}
}

@ARTICLE{ref39,
  author  = {W. Hua and Y. Wang and Z. Yang},
  title   = {Knowledge and data co-driven deep learning model for PolSAR image classification},
  journal = {Results Eng.},
  volume  = {29},
  year    = {2026},
  note    = {Art. no. 108947}
}

@ARTICLE{ref40,
  author  = {J. Geng and L. Dong and Y. Zhang and W. Jiang},
  title   = {Masked auto-encoding and scatter-decoupling transformer for polarimetric SAR image classification},
  journal = {Pattern Recognit.},
  volume  = {166},
  year    = {2025},
  note    = {Art. no. 111660}
}

@ARTICLE{ref41,
  author  = {L.-Y. Dai and M.-D. Li and S.-W. Chen},
  title   = {PolSAR Image Super-Resolution With Polarimetric Degradation Modulation Network},
  journal = {IEEE Trans. Aerosp. Electron. Syst.},
  volume  = {61},
  number  = {6},
  pages   = {17938--17954},
  year    = {2025}
}

@TECHREPORT{ref44,
  author      = {M. A. M. Vissers and J. J. van der Sanden},
  title       = {Groundtruth collection for the JPL-SAR and ERS-1 campaign in Flevoland and the Veluwe (NL) 1991},
  institution = {Landbouwuniversiteit Wageningen},
  address     = {Wageningen, The Netherlands},
  number      = {31},
  year        = {1992},
  note        = {68 pp.}
}

@INPROCEEDINGS{ref45,
  author    = {X. Liu and L. Jiao and F. Liu and D. Zhang and X. Tang},
  title     = {PolSF: PolSAR image datasets on San Francisco},
  booktitle = {Proc. Int. Conf. Intell. Sci.},
  address   = {Cham, Switzerland},
  publisher = {Springer},
  pages     = {214--219},
  year      = {2022}
}

@ARTICLE{ref46,
  author  = {S. Hochstuhl and N. Pfeffer and A. Thiele and S. Hinz and R. Scheiber and A. Reigber},
  title   = {Pol-InSAR-Island: A benchmark dataset for multi-frequency Pol-InSAR data land cover classification},
  journal = {ISPRS Open J. Photogramm. Remote Sens.},
  volume  = {3},
  year    = {2023},
  note    = {Art. no. 100047}
}

@ARTICLE{ref47,
  author  = {S. Quegan and T. {Le Toan} and J. Chave and J. Dall and J.-F. Exbrayat and D. H. T. Minh and M. Lomas and M. M. D'Alessandro and P. Paillou and K. Papathanassiou and F. Rocca and S. Saatchi and K. Scipal and H. Shugart and T. L. Smallman and M. W. J. Soja and S. Tebaldini and L. Ulander and L. Villard and M. Williams},
  title   = {The European Space Agency BIOMASS mission: Measuring forest above-ground biomass from space},
  journal = {Remote Sens. Environ.},
  volume  = {227},
  pages   = {44--60},
  year    = {2019}
}

@ARTICLE{ref48,
  author  = {L. Zhang and S. Zhang and B. Zou and H. Dong},
  title   = {Unsupervised deep representation learning and few-shot classification of PolSAR images},
  journal = {IEEE Trans. Geosci. Remote Sens.},
  volume  = {60},
  year    = {2022},
  note    = {Art. no. 5100316}
}

@ARTICLE{ref49,
  author  = {X. Tan and M. Li and P. Zhang and Y. Wu and W. Song},
  title   = {Complex-valued 3-D convolutional neural network for PolSAR image classification},
  journal = {IEEE Geosci. Remote Sens. Lett.},
  volume  = {17},
  number  = {6},
  pages   = {1022--1026},
  year    = {2019}
}

@ARTICLE{ref50,
  author  = {J. Ni and D. Xiang and Z. Lin and others},
  title   = {DNN-based PolSAR image classification on noisy labels},
  journal = {IEEE J. Sel. Topics Appl. Earth Observ. Remote Sens.},
  volume  = {15},
  pages   = {3697--3713},
  year    = {2022}
}
\end{document}